\documentclass[lettersize,journal]{IEEEtran}
\usepackage{booktabs}
\usepackage{amsmath,amsfonts}
\usepackage{algorithmic}
\usepackage{algorithm}
\usepackage{array}
\usepackage{subcaption}
\usepackage{textcomp}
\usepackage{stfloats}
\usepackage{url}
\usepackage{verbatim}
\usepackage{threeparttable}
\usepackage{graphicx}
\usepackage{cite}
\usepackage{graphicx}
\usepackage{wrapfig}  
\usepackage{booktabs}
\usepackage{enumitem}
\usepackage{subcaption}
\usepackage{adjustbox}  

\begin{document}
\bibliographystyle{unsrt}
\title{Improving Ensemble Forecasts of Abnormally Deflecting Tropical Cyclones with Fused Atmosphere-Ocean-Terrain Data}
\author{Qixiang Li,
        Yuan Zhou,~\IEEEmembership{Senior Member,~IEEE,}
        Shuwei Huo, 
        Chong Wang, 
        Xiaofeng Li,~\IEEEmembership{Fellow,~IEEE}
\thanks{Qixiang Li is with the School of Electrical and Information Engineering, Tianjin University, Tianjin 300072, China. (e-mail: liqixiang@tju.edu.cn)}
\thanks{Yuan Zhou is with the School of Electrical and Information Engineering, Tianjin University, Tianjin 300072, China (e-mail: zhouyuan@tju.edu.cn).}
\thanks{Shuwei Huo is with School of Computer Science and Engineering, Northeastern University, Shenyang 110819, China (e-mail: xiaofeng.li@ieee.org)}
\thanks{Chong Wang, Xiaofeng Li arewith the Key Laboratory of Ocean Circulation and Waves, Institute of Oceanology, and the Center for Ocean Mega-Science, Chinese Academy of Sciences, Qingdao 266071, China (e-mail: wangchong1@qdio.ac.cn, xiaofeng.li@ieee.org).}

}

\markboth{Journal of \LaTeX\ Class Files,~Vol.~14, No.~8, August~2021}%
{Shell \MakeLowercase{\textit{et al.}}: A Sample Article Using IEEEtran.cls for IEEE Journals}


\maketitle

\begin{abstract}
Deep learning-based tropical cyclone (TC) forecasting methods have demonstrated significant potential and application advantages, as they feature much lower computational cost and faster operation speed than numerical weather prediction models. However, existing deep learning methods still have key limitations: they can only process a single type of sequential trajectory data or homogeneous meteorological variables, and fail to achieve accurate forecasting of abnormal deflected TCs. To address these challenges, we present two groundbreaking contributions. First, we have constructed a multimodal and multi-source dataset named AOT-TCs for TC forecasting in the Northwest Pacific basin. As the first dataset of its kind, it innovatively integrates heterogeneous variables from the atmosphere, ocean, and land, thus obtaining a comprehensive and information-rich meteorological dataset. Second, based on the AOT-TCs dataset, we propose a forecasting model that can handle both normal and abnormally deflected TCs. This is the first TC forecasting model to adopt an explicit atmosphere-ocean-terrain coupling architecture, enabling it to effectively capture complex interactions across physical domains. Extensive experiments on all TC cases in the Northwest Pacific from 2017 to 2024 show that our model achieves state-of-the-art performance in TC forecasting: it not only significantly improves the forecasting accuracy of normal TCs but also breaks through the technical bottleneck in forecasting abnormally deflected TCs.
\end{abstract}


\section{Introduction}
\IEEEPARstart{A}{s} the impacts of extreme climate continue to intensify and a strong El Niño event emerges, the global climate system is entering a new phase of pronounced variability\cite{ripple2024StateClimate2024}. In recent years, multiple key climate indicators, including sea surface temperature and ice-sheet extent, have repeatedly reached record highs\cite{terhaar2025record}. Since April 2023, global sea surface temperatures have remained at historically elevated levels, with the average from April 2023 to March 2024 exceeding the previous record set in 2015–2016 by 0.25°C\cite{globalmarine2025}. This marked warming is reshaping global ocean circulation, vertical stratification, and the temperature–density structure of seawater\cite{marcos2025global}. For TCs, which derive their primary energy from ocean heat, elevated sea surface temperatures provide more favorable thermodynamic conditions for their genesis and development, and may further amplify the hazards they produce, including destructive winds, heavy rainfall, and storm surges\cite{wangAdvancingForecastingCapabilities2025,ma2026interactions}. Therefore, improving TC forecasting capability has become a major issue in disaster prevention and mitigation under the context of climate change. However, the evolution of TC track and intensity is jointly regulated by oceanic conditions, atmospheric circulation, and land-surface underlying processes, exhibiting pronounced nonlinear and coupled characteristics\cite{kangNorthEquatorialCurrent2024}. This continues to make accurate prediction of TC track and intensity extremely challenging.

At present, many countries still rely primarily on high-performance computing systems and Numerical Weather Prediction (NWP) models for TC forecasting. This class of methods is centered on General Circulation Models (GCMs), which predict relevant meteorological variables by parameterizing the fluxes of energy and matter across the land, ocean, and atmosphere. However, due to uncertainties in initial conditions, complex nonlinear physical processes, and high computational costs, NWP methods still suffer from insufficient accuracy and timeliness in representing the rapid evolution of atmospheric variables over short timescales (e.g., 6–12 hours).

With the continuous accumulation of multi-source meteorological observation data and the rapid advancement of deep learning techniques, the research paradigm in weather forecasting is gradually shifting from traditional Numerical Weather Prediction (NWP) toward data-driven models. Compared with NWP, deep learning methods offer computational efficiency advantages of several orders of magnitude, while their predictive accuracy has become comparable to that of traditional numerical schemes\cite{ruttgers2019prediction,huang2022mmstn}. At the same time, advances in remote sensing technology have significantly enhanced the capability to observe the ocean–atmosphere–land systems associated with TCs\cite{hu2020comparing,chen2020novel,tian2021estimation}. Multi-source remote sensing products, such as satellite-retrieved sea surface temperature (SST), sea surface salinity (SSS), and digital elevation models (DEM), can characterize the dynamic interactions among the ocean, land, and atmosphere from different perspectives, and have already demonstrated substantial potential in TC monitoring and prediction\cite{wang2021tropical,jin2024towards,zhang2019tropical,ma2024multiscale}. 

\begin{figure*}[t]
    \centering
    \includegraphics[width=0.9\textwidth]{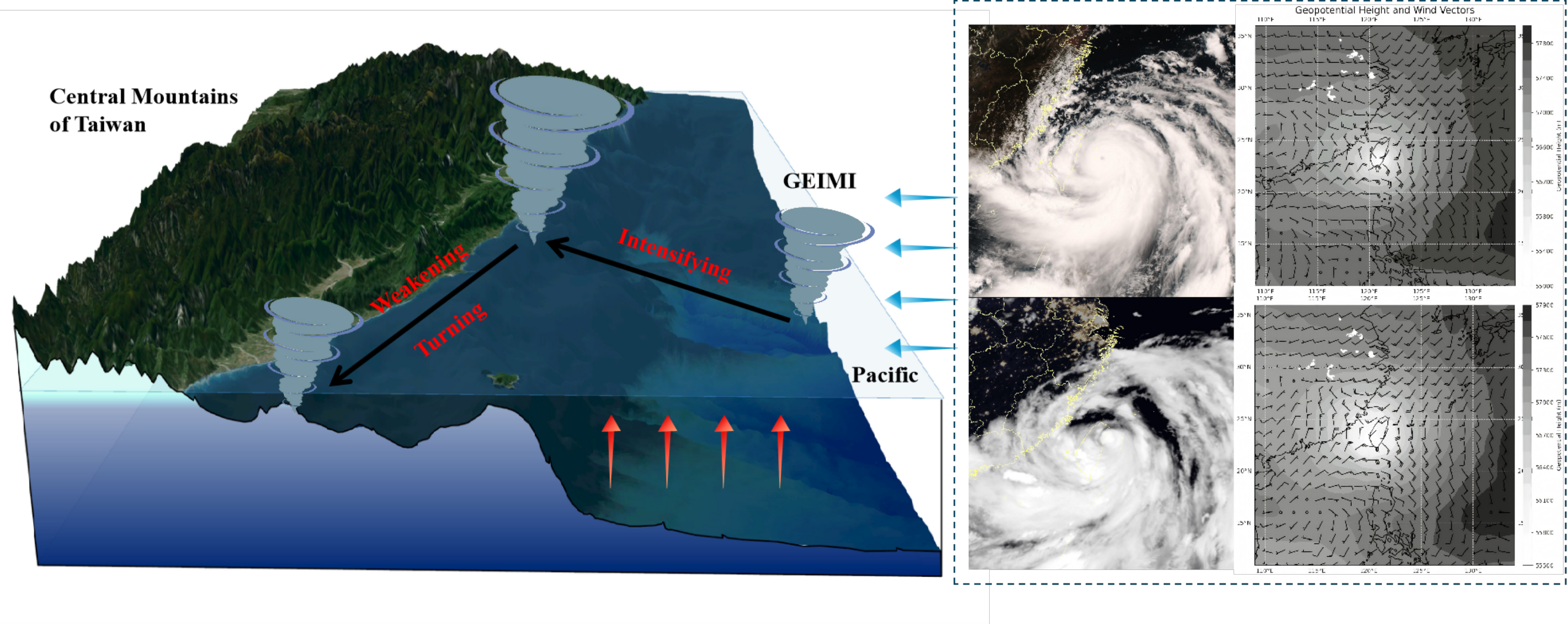}
    \caption{Abnormal deflection of TC GEIMI during landfall in Taiwan in 2024. The Central Mountain Range (peaking at nearly 4,000 m) interacted with GEIMI's low-level circulation within the 600–700 hPa vertical layer, inducing a venturi effect that guided its low-level center southward. Satellite imagery revealed complete eye dissolution after GEIMI's collision with the mountain range.} 
    \label{fig:1} 
\end{figure*}

Although existing studies have attempted to incorporate multimodal data into TC forecasting, prediction of TCs with anomalous tracks remains insufficient. Compared with samples following typical tracks, anomalous-track typhoons are more susceptible to the combined influences of large-scale atmospheric circulation adjustments, changes in the oceanic background state, and the blocking and steering effects of underlying surface topography. As a result, their track evolution usually exhibits stronger abruptness, discontinuity, and uncertainty. As shown in Figure \ref{fig:1}, the complex nonlinear coupling among oceanic, terrestrial, and atmospheric processes further increases the difficulty of track prediction. In the Western North Pacific (WNP), in particular, TC genesis locations and track evolution are jointly regulated by the seasonal configuration of atmospheric circulation and the state of the ocean. Recent studies have shown that multiple canonical track patterns exist in this region \cite{fuLongtermTrendTropical2025}, and that these patterns are not continuously distributed, but are instead composed of several mutually separated track sub-modes\cite{ZhuAttention2025}. However, existing methods face two major limitations: on the one hand, they struggle to effectively integrate heterogeneous information from reanalysis data and remote sensing data; on the other hand, most rely on fixed or implicit assumptions of continuous distributions, making them prone to domination by mainstream sample patterns. This often causes anomalous-track samples to be “averaged out” \cite{baiman2026watch} and may even lead to mode collapse\cite{hoangMGANTRAININGGENERATIVE2018,dendorferMGGANMultiGeneratorModel2021}. Consequently, current methods have difficulty stably characterizing the distributional features and evolutionary patterns of anomalous-track TCs.

Based on the above observations, this paper proposes AOT-TCNet, a multimodal TC forecasting model based on ocean–land–atmosphere coupling. To address the two core challenges in anomalous-track prediction, namely the strong heterogeneity of driving factors and the pronounced multimodality of track distributions, AOT-TCNet first employs a set of specially designed encoders to separately extract feature representations from oceanic, land-surface, and atmospheric processes. This design helps preserve the physical semantics and scale-specific information of each modality. This design preserves the physical semantics and scale information of different modalities. A cross-modal fusion mechanism is then introduced to enhance the model’s ability to jointly represent key ocean–atmosphere–land driving signals, thereby enabling more comprehensive identification of the environmental background associated with the formation of anomalous tracks.

Building on this foundation, AOT-TCNet further introduces TMA-MoE, a mode-adaptive mixture-of-experts system, which decomposes the complex distribution of TC tracks into several track sub-modes. This allows different experts to focus on different track patterns and their evolution, thereby reducing the masking effect of conventional-track samples on anomalous-track samples and alleviating the problem of mode collapse in continuous latent space learning. As a result, the proposed method can characterize multimodal TC track distributions more stably and improve the predictive consistency, robustness, and confidence for anomalous TCs, thereby providing a more reliable solution for typhoon track forecasting under complex ocean–atmosphere conditions.

In summary, the main contributions of this study are as follows:

\begin{itemize}[leftmargin=10pt]
\item We pioneer the incorporation of complex underlying terrain elevation under TC conditions into a multimodal deep learning framework, which enables explicit quantification of physical mechanisms such as terrain-lifted wind acceleration and anomalous north–south deflection, effectively addressing the forecasting challenges for abnormally deflected TCs.

\item We introduce TMA-MoE, which effectively captures diverse TC motion patterns and significantly improves the model’s adaptability, robustness, and predictive accuracy.

\item We construct the most comprehensive TC dataset to date for the WNP region, with the longest temporal coverage currently available, and integrate environmental variables from multiple physical domains, including the atmosphere, ocean, and land.
\end{itemize}

Extensive experiments conducted on all TC cases from 2017 to 2024 demonstrate that AOT-TCNet achieves state-of-the-art performance in TC forecasting tasks, with particularly strong capability in predicting anomalously deflected TC cases that have traditionally been considered difficult to forecast accurately. Notably, several evaluation metrics even surpass the performance of operational official forecasting systems.

\section{RELATED WORK}
\label{gen_inst}
TC forecasting has long been recognized as one of the central challenges in meteorology. Conventional TC prediction methods primarily rely on NWP models; however, such models are typically computationally expensive and exhibit limited inference efficiency, making them difficult to deploy for fine-grained and near–real-time operational applications. In recent years, deep learning approaches have achieved breakthrough progress in weather forecasting owing to their adaptive representation of complex nonlinear spatiotemporal patterns and their markedly superior inference efficiency compared with traditional models \cite{chenFuXiCascadeMachine2023,bi2023accurate,lam2023learning}. As a result, they have gradually emerged as a key research direction in intelligent meteorological forecasting. In TC track prediction, early studies largely drew inspiration from trajectory prediction research \cite{alahi2016social}, employing recurrent neural networks (RNNs) and their variants to model individual TC track sequences \cite{moradi2016sparse,alemany2019GBRNN,pan2019DLM,gao2018nmpt}. However, due to the strong coupling between large-scale circulation backgrounds and mesoscale to synoptic-scale weather systems, TC tracks exhibit pronounced nonlinearity and diversity. Under extreme meteorological conditions, it is often difficult to characterize future TC evolution using a single deterministic track, and substantial challenges remain, particularly for forecasting anomalously turning TCs. Motivated by these limitations, recent studies have increasingly introduced generative modeling paradigms and multimodal meteorological data into TC forecasting tasks.

\subsection{Generative Model–Based TC Forecasting}
Generative models have been systematically studied in the trajectory prediction domain \cite{kosaraju2019social,amirian2019social,dendorfer2020goal,fernando2018gd}, with the core objective of modeling the probability distribution of future trajectory states, thereby providing a theoretical foundation for uncertainty modeling and multi-solution prediction. Inspired by this paradigm, Rüttgers et al. \cite{ruttgers2019prediction} were the first to introduce generative adversarial networks (GANs) into TC track prediction and demonstrated the feasibility and potential of deep learning for TC forecasting using satellite remote sensing imagery. Subsequently, Huang et al. \cite{huang2022mmstn} proposed the multimodal trajectory prediction network MMSTN, which generates multiple plausible future tracks via a GAN-based architecture, effectively enhancing the representation of track uncertainty. Building upon this work, they further developed efficient utilization strategies for multi-source meteorological data \cite{huangMGTCFMultiGeneratorTropical2023}, validating the advantages of deep generative models in capturing complex atmospheric structures and evolution patterns.

Despite the strong capability of GANs in modeling multimodal trajectory distributions, most existing methods still rely on supervised learning paradigms and use a single ground-truth trajectory as the training target \cite{subich2025fixing}. This setting inherently limits the model’s ability to fully learn multimodal distributions and makes it difficult to fundamentally alleviate the issue of mode collapse \cite{gupta2018SGAN,sadeghian2019sophie,goodfellow2014generative}. To address this challenge, mixture-of-experts systems decompose the generative task into multiple subproblems, with different experts specializing in distinct data modes, thereby preventing a single generator from converging to a limited set of modes. Based on this insight, our approach introduces the TMA-MoE (Tropical Cyclone Mode-Adaptive Mixture of Experts) generative modeling framework to enhance the multimodal representational capacity and physical consistency of TC track prediction.

\subsection{Multimodal Datasets for TC Forecasting}
As deep learning models continue to advance in their ability to fuse information from multiple sources, multimodal fusion architectures have attracted growing attention in TC forecasting research. Correspondingly, a variety of multimodal datasets tailored for TC prediction have been developed. The deep integration of deep learning with multi-source meteorological data not only substantially improves forecasting accuracy but also provides a data foundation for building highly reliable and robust intelligent TC forecasting models.

Existing TC datasets typically center spatial samples on the TC core location and select spatial windows that cover sufficient environmental context, thereby jointly characterizing TC internal structure and the surrounding large-scale environmental fields across multiple scales. On a global scale, the Hurricane Satellite (HURSAT) dataset released by NOAA is one of the most widely used and influential benchmark datasets in TC research \cite{knapp2017hursat}. It provides long-term satellite observations spanning 1978–2015 and has served as a critical data foundation for studies on TC structural evolution, intensity variation, and statistical analysis. Building upon HURSAT, the TCIR dataset proposed by Chen et al. \cite{chen2018tcir} further reorganized and refined the samples and introduced multi-channel radiance information, significantly improving data suitability for deep learning models. Huang et al. \cite{huang2025benchmark} subsequently proposed TCND, an open multimodal TC dataset covering six ocean basins over a 70-year period, which systematically integrates intrinsic TC attributes, meteorological gridded data, and environmental variables, providing comprehensive support for deep learning models to understand TC motion mechanisms. Wang et al.\cite{wang2025global} introduced the SETCD dataset, which extracts TC-centered multivariable regions from the GridSat-B1 and ERA5 datasets, including infrared, water vapor, and visible channels from GridSat-B1, as well as 69 meteorological variables from ERA5. For the Western North Pacific basin, the Digital TC dataset \cite{kitamoto2023digital} provides longer temporal coverage and higher spatiotemporal resolution satellite observations, offering critical support for systematic investigations of TC genesis, development, and evolution in this region. 

Based on these multimodal datasets, existing TC forecasting methods have been able to capture atmosphere–ocean interaction processes to a certain extent \cite{giffard2020FFN,wangAdvancingForecastingCapabilities2025}. However, most prior studies have largely overlooked the critical role of topographic constraints in TC evolution, including buffering effects induced by continuous terrain blocking and the Venturi effect triggered by terrain-induced uplift \cite{ma2025philippine}. These observations indicate the necessity of further advancing atmosphere–ocean–topography coupled modeling frameworks to improve the completeness and accuracy of TC forecasting under complex underlying surface conditions. Consequently, constructing a new multimodal TC dataset that simultaneously integrates atmospheric, oceanic, and topographic information is a crucial prerequisite for achieving this objective.

\section{Dataset}
We collected atmospheric, terrain, and oceanic datasets related to TCs from 1950 to 2024, which are closely associated with TC genesis and decay. This comprehensive dataset allows for a holistic exploration of TCs, enabling the revelation of their evolutionary patterns and characteristics throughout their lifecycle.

{\bf{CMA-BST Dataset.}} TC track and intensity characteristics were obtained from the \textit{China Meteorological Administration Tropical Cyclone Best Track Dataset(CMA-BST)}, which records key information for each TC lifecycle in the Western North Pacific and South China Sea, including longitude, latitude, minimum central pressure, and 2-minute average maximum near-center wind speed.

{\bf{ERA5 Reanalysis Data.}} The \textit{ERA5 reanalysis dataset}, provided by the European Centre for Medium-Range Weather Forecasts (ECMWF), offers a spatial resolution of 0.25°×0.25° with 6-hourly updates. We utilized wind fields and geopotential height to characterize TC atmospheric circulation and pressure structures. Specifically, centered on the current cyclone position, we extracted the u-wind, v-wind, and geopotential height within a 25°×25° spatial window around the TC center. Inspired by statistical prediction models and sensitivity analyses, and to capture air movements in the lower, middle, and upper troposphere, pressure levels at 200, 500, 850, and 1000 hPa were selected.

{\bf{Global Ocean Science Database (CODCv1).}} The CODCv1 in-situ ocean observation dataset, provided by the CAS-Ocean Data Center, was employed to analyze TC intensity, which is regulated by energy exchange between the ocean and atmosphere—particularly surface evaporation flux. Recent studies have demonstrated that the upper-ocean salinity barrier layer (BL) induces specific upper-ocean responses to Western North Pacific TCs: during TC passage, surface salinity significantly (SSS) increases on both sides of the track, while SST slightly decreases. The BL limits TC-induced SST cooling by suppressing cold water entrainment from the thermocline. Therefore, SST and salinity were selected as predictors for TC intensity changes.

{\bf{GEBCO Dataset.}} The Global Bathymetric Dataset from the Intergovernmental Oceanographic Commission (IOC) and International Hydrographic Organization (IHO) provides high-precision global seafloor topography data from coastlines to deep oceans, including detailed representations of mid-ocean ridges, trenches, and plateaus. 

In addition to conventional gridded data, we incorporated critical non-spatial variables to enhance the dataset's multidimensional information. These variables include the translational speed of TC centers, historical movement direction, historical intensity variability, month, and the Niño3.4 index. Historical movement direction and intensity variability provide essential prior information for TC trajectory and intensity forecasting; monthly data reflect seasonal cyclicity characteristics, with significant differences in formation and movement patterns of western Pacific TCs across seasons; translational speed is intrinsically linked to TC lifecycle dynamics. Furthermore, recent studies demonstrate that average TC genesis locations and trajectory modes exhibit significant ENSO (El Niño-Southern Oscillation) covariability \cite{fuLongtermTrendTropical2025}, with characteristic oceanic thermal response patterns observed during TC development stages. By integrating these heterogeneous data sources, the \textit{AOT-TCs} dataset enriches TC prediction models with enhanced informational dimensionality, thereby improving forecast accuracy and advancing understanding of complex physical processes.

In summary, we present \textit{AOT-TCs}, the longest and most comprehensive TC dataset for the Northwest Pacific, covering the period from 1950 to 2024. Unlike existing public datasets, \textit{AOT-TCs} integrates multimodal information across diverse physical variables influencing TC dynamics rather than relying on single-sensor observations, thereby establishing a unified framework for TC prediction research. 

\section{METHOD}
TC forecasting aims to predict future trajectories and intensity of TCs given historical states. Specifically, we take the time series $X=\left\{x_{\text {lon }}, x_{\text {lat }}, x_{\text {traj }}, x_{\text {int }}\right\}$ consisting of longitude, latitude, central pressure, and wind speed as input. For a given historical sequence $\left\{X_{\text {-t+1 }}, X_{\text {-t+2 }},... X_{\text {0 }}\right\}$ with fixed temporal length, the objective is to predict the future trajectory and intensity , where each trajectory contains the subsequent $\hat{Y}=\left\{\hat{Y}_{\text {lon }}, \hat{Y}_{\text {lat }}, \hat{Y}_{\text {traj }}, \hat{Y}_{\text {int }}\right\}$ time steps of positional and intensity evolution. 

\begin{figure*}[t]
    \centering
    \includegraphics[width=\textwidth]{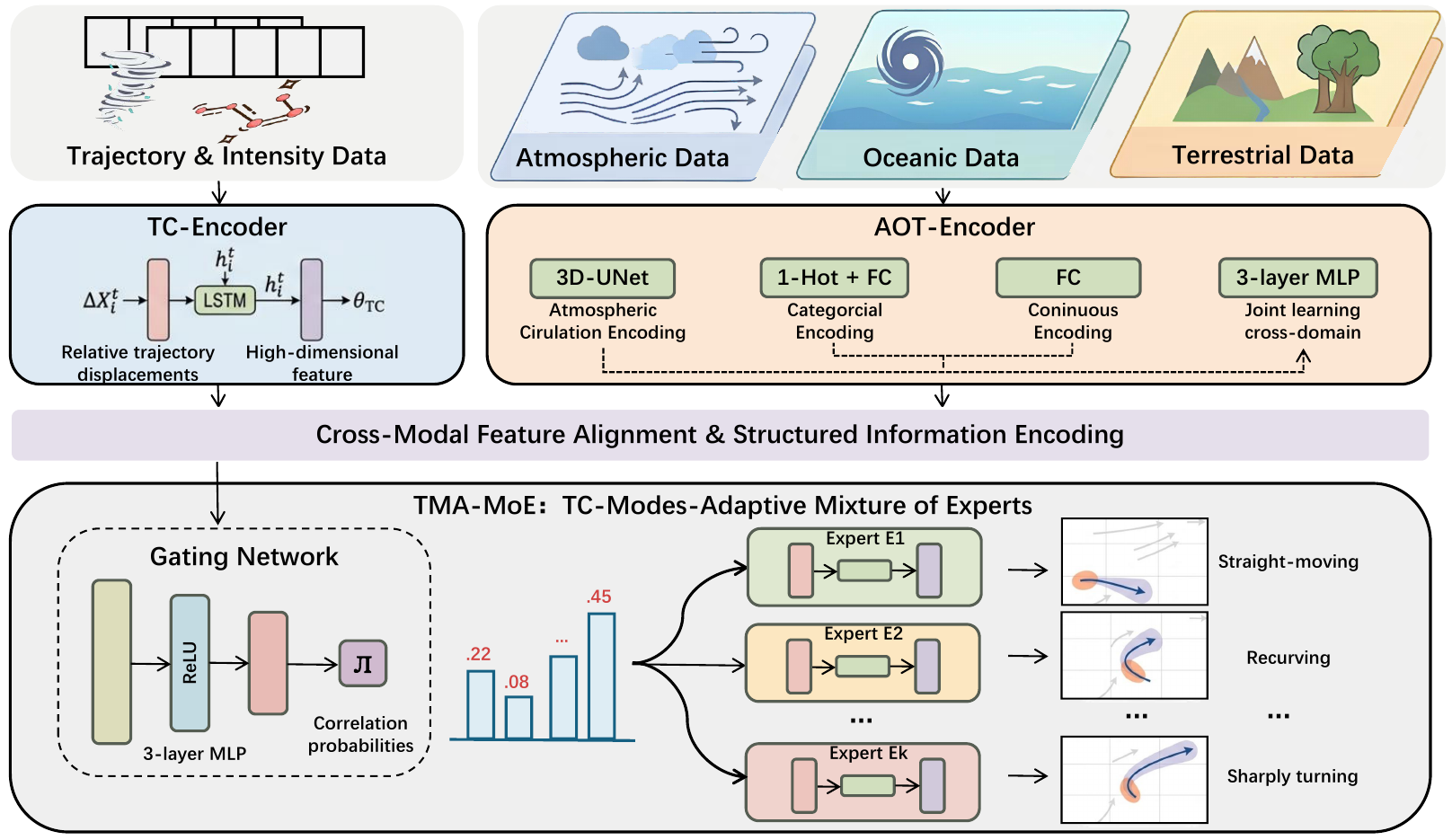} 
    \caption{The architecture of AOT-TCNet, comprising Encoders and TMA-MoE. The environment data $D$ and observed trajectories $X$ are encoded and passed to the fusion module. The experts can predict different mode trajectory distributions for the given observation. The gating network estimates probabilities $\pi$ for the experts. The model samples or selects a expert from $\pi$ and predicts a trajectory $Y$ conditioned on the features $c$.}
    \label{fig:3} 
\end{figure*}

\subsection{Encoders} 
The overall architecture of the proposed model is illustrated in Fig. \ref{fig:3}. We first employ a layered embedding strategy to independently encode TC, oceanic, terrain, and atmospheric variables, enabling the extraction of key dynamic features from each physical domain. These multi-source features are then projected into a unified representation space through a cross-modal feature alignment mechanism and subsequently concatenated into a fused vector $c$. The fused representation is further processed by a two-layer MLP, allowing the network to explicitly learn cross-domain dependencies among atmospheric circulation, ocean heat content, and underlying surface conditions. This design facilitates the capture of the co-evolutionary mechanisms governing TC development and enhances both the prediction accuracy and generalization capability of the model.


{\bf{TC-Encoder.}} For TC sequence samples, we utilize LSTM to encode historical trajectories and extract dynamic features. The TC $x_{\text{i}}$ are encoded into high-dimensional features $\theta _{\text{TC}}$ through the following formulation:
\begin{align}
c_{\text{TC}} = \text{LSTM}(x_{\text{i}}^{t}, h_{\text{i}}^{t})
\end{align}
Where $h_{\text{i}}^{t}$ denotes the hidden state of the LSTM at time step $t$.

{\bf{AOT-Encoder.}} To efficiently integrate multimodal data from different physical domains into TC forecasting models and deeply explore their distinct physical processes and interaction mechanisms, we perform multi-scale feature extraction and fusion on ocean, terrestrial, and atmospheric data to construct a unified multimodal representation that is precisely embedded into existing TC prediction model architectures. 
First, each category in $D_{\text{cat}}$ is converted to one-hot encoding and then processed through a two-layer MLP to create categorical embedding $c_{\text{cat}}$ . Similarly, $D_{\text{cont}}$ is encoded into continuous embedding $c_{\text{cont}}$ using a two-layer MLP. $D_{\text{atm}}$ employs CNN networks to extract spatiotemporal features from historical atmospheric circulation data, encoding them into atmospheric embedding $c_{\text{atm}}$ :
\begin{align}
c_{\text{atm}} = \text{3D-UNet}(D_{atm}(z_{\text{i}}, u_{\text{i}},v_{\text{i}}))
\end{align}
where $z$, $u$, $v$ represent geopotential height, meridional wind, and zonal wind respectively.
Finally, these features are concatenated into $c=c_{TC}+c_{\text{atm}}+c_\text{cat}+c_{\text{cont}}$ , where ``$+$'' denotes vector concatenation operation. The concatenated features are input through FC, enabling the model to simultaneously learn and capture correlations across ocean, terrestrial, and atmospheric domains. This module design allows the model to comprehensively consider atmospheric circulation, ocean heat content, underlying surface variations, and other key factors, thereby improving the physical consistency and generalization capability of TC predictions.

\subsection{TC-Modes-Adaptive Mixture of Experts (TMA-MoE)}
Mixture of Experts (MoE) enhances model capacity by replicating network modules into multiple expert submodels \cite{jacobs1991textordfeminineadaptive}. Sparse MoE (SMoE) \cite{zhu2024moe,zhu2023sira} further introduces a gating mechanism that activates only a small subset of experts, significantly reducing computational overhead while preserving expressive power, and has been widely adopted in multi-task learning and model compression \cite{li2023merge,chen2023adamv}.

In recent years, MoE has been incorporated into Transformer architectures to replace dense layers, thereby improving the performance of large language models \cite{shazeer2017outrageously,chen2022towards,zhong2022meta}. By enabling specialization among experts, MoE allows models to outperform dense counterparts of comparable scale in multi-domain and multimodal tasks. In principle, the gating network routes inputs with similar patterns to specific experts while suppressing irrelevant features, enabling efficient and targeted representation learning \cite{chowdhury2023patch}.

This property offers a promising paradigm for TC forecasting in meteorology. Accordingly, we propose a mode-adaptive mixture-of-experts framework that can accommodate diverse TC patterns. Specifically, a router network selectively activates experts corresponding to different modes, treating the multimodal target distribution as a mixture of continuous trajectory distributions, thereby enabling the learning of distinct categories of TC behaviors.
Recent advances further enhance large language models (LLMs) by replacing dense layers in Transformers with MoE structures. 
The core idea is to construct multiple specialized expert submodels, each trained to excel in specific domains such as text generation or logical reasoning. 

Specifically, instead of approximating the data distribution with a single distribution as in conventional approaches, we employ a mixture of multiple distributions and explicitly enlarge the divergence among them so that each component captures distinct TC modes. Each expert $E_k$ maps the conditional feature $c$ to the output space $Y$, thereby inducing an individual distribution $P_k$. Collectively, the $k$ experts define a mixture distribution in the data space, denoted as $P_{\text{model}}$, which is composed of these $k$ component distributions.

{\bf{Router Network.}} To select the most suitable expert for each TC sample, we design a learnable router network. 
It takes encoded features as input and outputs the relevance probabilities between the input and all experts.

Formally,
\begin{equation}
\mathbf{W} = \text{softmax}(G_{\theta}(c)),
\end{equation}
where $G_{\theta}$ denotes the router network parameterized by $\theta$, implemented as a 3-layer MLP with ReLU activation and hidden dimension of 48. 
$\mathbf{W}$ represents routing weights assigning sample $x$ to each expert.

We adopt a hard routing strategy by selecting the top-$k$ weights, and in this work, we set $k=1$.

To enable adaptive expert selection under different environmental conditions and TC dynamics, we introduce a probabilistic modeling approach for routing decisions. 
Instead of directly modeling generator likelihood, we measure the matching quality between predicted trajectories and ground truth.

Let $\hat{Y}_i$ denote the trajectory predicted by the $i$-th expert and $Y$ the ground truth. 
The matching score is defined as:
\begin{equation}
S_i = - \| \hat{Y}_i - Y \|^2,
\end{equation}
where a smaller prediction error leads to a higher score.

The scores are normalized and combined with Bayes' rule to obtain the posterior distribution over experts:
\begin{equation}
R_i = \frac{\exp(S_i)}{\sum_{j=1}^{k} \exp(S_j)}.
\end{equation}

Finally, the router network outputs a probability distribution $\pi(c_i)=[\pi_1(c_i),\pi_2(c_i),...,\pi_k(c_i),]$ over expert selection, with the objective of approximating the data-driven posterior distribution $R_i$. To this end, we optimize the routing network using a routing entropy–based objective:
\begin{equation}
\mathcal{L}_{R_{Entropy}} = - \sum_{i=1}^{k} R_i \log \pi_i.
\end{equation}

{\bf{Modes-Adaptive Experts.}} We reconstruct multiple independent generative networks into an MoE framework. 
Instead of modifying standard linear or convolutional layers, each expert is designed to learn a specific TC trajectory mode. 
The mixture of multimodal distributions approximates the real data distribution while encouraging diversity among experts.

All expert networks share the same architecture but do not share weights. 
Given encoded feature $c$, only selected experts are activated.
Let the expert set be:
\begin{equation}
E = (E_1, E_2, ..., E_k),
\end{equation}

Here, $E_g(\cdot)$ represents the generation function of the $g$-th expert.

Through training, the router learns to select the expert that generates trajectories best matching the current TC path mode, enabling adaptive modeling of diverse TC behaviors.

\subsection{Training Loss}
In this subsection, we describe the loss functions used in the proposed framework in detail. To effectively capture the multimodality and uncertainty in TC trajectory evolution, we formulate the training objective as a joint optimization problem composed of several complementary loss terms, including the reconstruction loss, distribution contrastive loss, and mode classification loss. These loss functions not only ensure prediction accuracy, but also guide the model to learn a reasonable trajectory distribution and explicitly encourage mode differentiation among multiple experts.

{\bf{Reconstruction Loss.}}
The reconstruction loss directly supervises the sample-level error between the generated results and the target results. For the TC sequence data generated by each expert, we adopt the Huber loss for optimization, which maintains high sensitivity in the small-error region while improving robustness in the presence of outliers.
\begin{align}
L_{\text{traj},i}^{\text{huber}} = 
\begin{cases} 
\frac{1}{2}(Y_i^t - \hat{Y}_i^t)^2, & \text{if } |Y_i^t - \hat{Y}_i^t| < \delta \\
\delta |Y_i^t - \hat{Y}_i^t| - \frac{1}{2}\delta^2, & \text{otherwise}
\end{cases}
\end{align}
Where $\delta$ is the threshold, which is set to 0.5 in this paper.

To supervise the numerical accuracy of meteorological variables, we employ the mean squared error (MSE) loss. This loss directly constrains the ability of the atmospheric encoder to reconstruct the intensity and spatial structure of meteorological variables, thereby providing physically consistent meteorological prior features for the downstream experts:
\begin{align}
L_{a,i}^{\text{MSE}} = \frac{1}{N} \sum_{i=1}^N (\hat{Y}_i^{atm} - Y_i^{atm})^2
\end{align}
{\bf{Disribution Contrastive Loss.}}
The primary goal of the multi-expert generation framework is to make the generated samples as close as possible to the real data distribution at the global level. To this end, we introduce a distribution contrastive loss to measure whether the input samples remain consistent with the manifold of the target data. For a real sample $Y\sim P_(data)$ and a generated sample $Y_e=E_k(c)$, the optimization objective of the expert system is defined as:
\begin{align}
L_D = -\mathbb{E}_{Y \sim p_{\text{data}}} [\log D(Y)] - \mathbb{E}_{c \sim p(c)} [\log(1 - D(E_k(c)))]
\end{align}
During training, this loss drives the predicted trajectory distribution to gradually approach the low-dimensional manifold where real TC trajectories lie, thereby effectively capturing the high-order structural characteristics of TC trajectories.

{\bf{Mode Classification Loss.}}
Using only the reconstruction loss and the distribution matching loss can ensure that the generated results globally approach the real distribution and that multiple experts receive relatively balanced training. However, this does not necessarily mean that different experts will learn different TC patterns. To explicitly encourage expert specialization, we introduce a mode classification loss, whose purpose is to determine from which expert branch a generated sample comes. For a sample generated by the k-th expert, the classifier outputs a k-dimensional probability vector $C(Y_i)=[C_1(Y_i),C_2(Y_i),...,C_k(Y_i)]$, where each element represents the predicted probability that sample $Y_i$ is generated by the i-th expert. Based on the above definition, the mode assignment classification loss can be expressed as:
\begin{align}
L_{\text{cls}} = \mathbb{E}_{c \sim p(c)}[-\log C_k(E_k(c))]
\end{align}
The mode assignment classification loss helps organize the overall generated distribution into several submanifolds handled by different experts. This encourages the model to learn TC forecasting experts with clear semantic distinctions, such as straight-moving, recurving, or sharply turning TC track patterns. The total training loss is the combination of all loss terms.

The final objective function is equivalent to:
\begin{align}
\min_E \left( \operatorname{JSD}(P_{\text{data}} \| P_{\text{model}}) - \beta \cdot \operatorname{JSD}_{\pi}(P_{E_1}, P_{E_2}, \dots, P_{E_k}) \right)
\end{align}
The first term corresponds to the objective of TMA-MoE, which aims to make the mixture distribution of the forecasts $P_{model}$ approximate the true data distribution $P_{data}$; the second term is the $JSD$ among the experts, which encourages the distributions of different experts to be far apart during optimization, preventing them from collapsing into the same mode.

\section{RESULTS AND DISCUSSION}
\subsection{Experimental Settings}
{\bf{Implementation Details.}} We conduct experiments using an NVIDIA GTX 4090D GPU. The model is optimized with the Adam optimizer, configured with an initial learning rate of 0.0001, a batch size of 128, and trained for 300 epochs. In our experiments, we input historical TC trajectory data from the past 48 hours (8 timesteps) to predict future TC trends for the next 24 hours (4 timesteps).

{\bf{Evaluation Metrics.}} We adopt the same evaluation configuration as before\cite{dendorferMGGANMultiGeneratorModel2021}. For path forecasting, we calculate the absolute distance between the actual position and the forecast position, and evaluate the average position error between the predicted trajectory and the ground truth (GT) trajectory. Notably, the error here is the Haversine distance (km) between the Earth's longitude and latitude coordinates. For intensity forecasting, we calculate the mean absolute error between the actual air pressure (hPa) and wind speed (m/s) and the forecasted air pressure and wind speed. Considering the inherent uncertainty of the future and the uncertainty of TC movement, we generate ensemble forecasts for each past trajectory, which is the same as the prediction by meteorological agencies.



\subsection{Comparison With State-of-the-Arts}
We evaluated our method using data from 2017 to 2019 as the test set and conducted comprehensive comparisons against existing techniques, including pedestrian trajectory prediction models, state-of-the-art TC forecasting approaches, and forecasts issued by the China Meteorological Observatory (CMO). As shown in Table \ref{tab:1}, our approach delivered the most promising results, particularly outperforming the CMO in ultra-short-term forecasting scenarios. Specifically, when forecasting 6, 12, and 18 hours ahead, our trajectory prediction errors were 13 km, 10.95 km, and 1.11 km lower than those of the CMO, respectively. However, at the 24-hour mark, the CMO outperformed our method, and as the forecast horizon extended, the performance gap between our model and traditional NWP methods gradually narrowed, eventually favoring the latter.

\begin{table*}[!htb]
    \centering
    \caption{Comparisons of average absolute error of TC prediction of different methods}
    \begin{tabular*}{\textwidth}{@{\extracolsep{\fill}}c*{12}{c}@{}}
        \toprule
        & \multicolumn{4}{c}{Distance (km)} & \multicolumn{4}{c}{Pressure (hpa)} & \multicolumn{4}{c}{Wind (m/s)} \\
        \cmidrule(lr){2-5} \cmidrule(lr){6-9} \cmidrule(lr){10-13}
        Methods & 6h & 12h & 18h & 24h & 6h & 12h & 18h & 24h & 6h & 12h & 18h & 24h \\
        \midrule
        SGAN\cite{gupta2018SGAN} & 28.88 & 61.75 & 98.74 & 140.61 & 1.91 & 3.12 & 4.2 & 5.12 & 1.05 & 1.69 & 2.28 & 2.81 \\
        GBRNN\cite{alemany2019GBRNN} & 29.93 & 65.06 & 105.74 & 152.06 & - & - & - & - & 1.16 & 1.89 & 2.52 & 3.1 \\
        MMSTN\cite{huang2022mmstn} & 27.57 & 59.09 & 96.54 & 139.19 & 1.69 & 2.86 & 3.94 & 4.74 & 0.95 & 1.52 & 2.1 & 2.55 \\
        MGTCF\cite{huangMGTCFMultiGeneratorTropical2023} & 23.14 & \underline{43.37} & 67.09 & 93.08 & \underline{1.37} & \underline{2.04} & \underline{2.66} & \underline{3.29} & 0.73 & 1.17 & 1.55 & 1.86 \\
        TCN$_m$\cite{huang2025benchmark} & \textbf{22.98} & 43.83 & 66.41 & 93.76 & - & - & - & - & \textbf{0.7} & \underline{1.09} & \underline{1.43} & \underline{1.75} \\
        CMO\cite{CMO} & 37.08 & 52.93 & \underline{60.69} & \textbf{75.49} & 2.67 & 4.3 & 5.04 & 6.31 & 2.29 & 3.45 & 2.75 & 5 \\
        Our & \underline{24.03} & \textbf{41.98} & \textbf{59.58} & \underline{86.05} & \textbf{1.22} & \textbf{1.74} & \textbf{2.19} & \textbf{2.58} & \underline{0.71} & \textbf{1.03} & \textbf{1.23} & \textbf{1.41} \\
        \bottomrule
    \end{tabular*}
    \begin{tablenotes}
        \item *Note: The implementations of FFN, BiGRU-att and DBF-Net are not available, we compare against their published 2014-2017 results.
    \end{tablenotes}
    \label{tab:1}
\end{table*}

\begin{figure*}[!htb]
    \centering
    \includegraphics[width=1.0\textwidth]{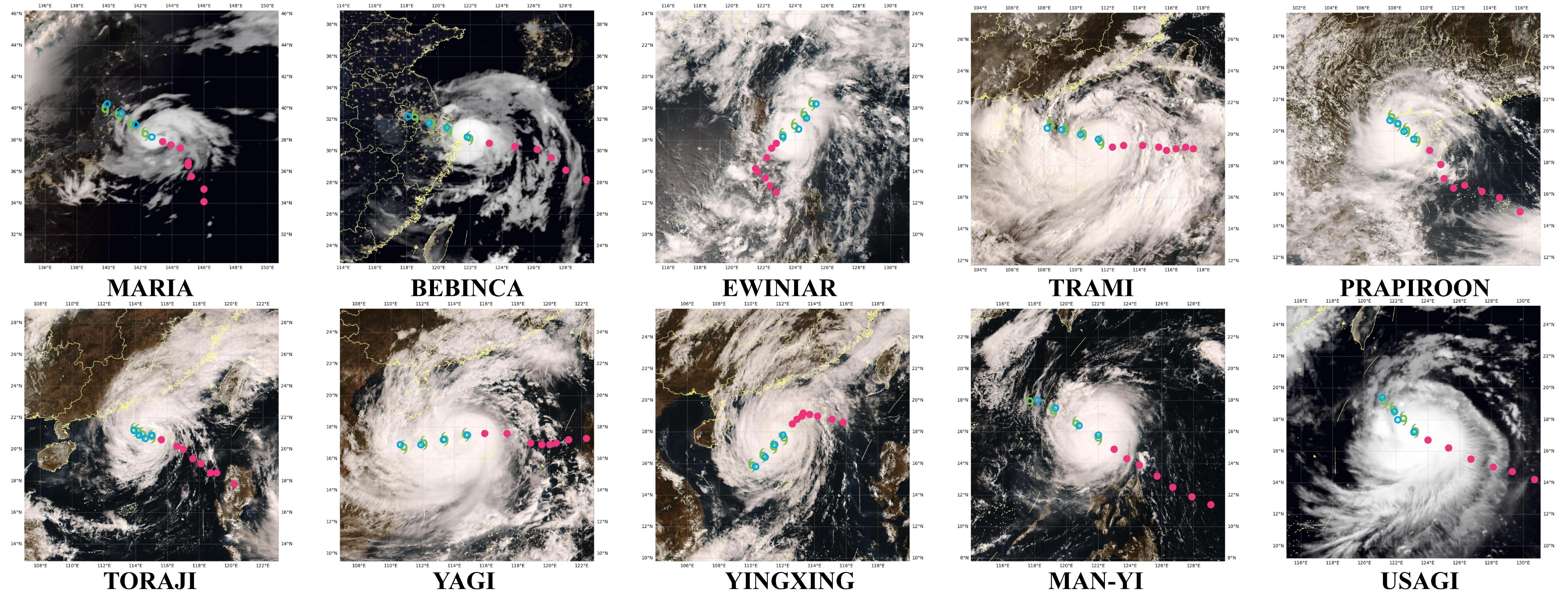} 
    \caption{Forecast results for selected TCs in 2024. Historical trajectories are shown in red, ground truth future paths in blue, and model-predicted trajectories in green.} 
    \label{fig:5} 
\end{figure*}

\begin{table*}[t]
    \centering
    \caption{Comparison of different methods for TC prediction from 2020 to 2024}
    \begin{tabular*}{\textwidth}{@{\extracolsep{\fill}}c*{12}{c}@{}}
        \toprule
        & \multicolumn{4}{c}{Distance (km)} & \multicolumn{4}{c}{Pressure (hpa)} & \multicolumn{4}{c}{Wind (m/s)} \\
        \cmidrule(lr){2-5} \cmidrule(lr){6-9} \cmidrule(lr){10-13}
        Methods & 6h & 12h & 18h & 24h & 6h & 12h & 18h & 24h & 6h & 12h & 18h & 24h \\
        \midrule
        MMSTN & 28.47 & 59.53 & 94.82 & 135.2 & 2.05 & 3.58 & 5.01 & 6.14 & 1.08 & 1.83 & 2.54 & 3.12 \\
        MGTCF & \textbf{25.21} & \underline{46.63} & \underline{71.04} & \underline{98.74} & \underline{1.57} & \underline{2.51} & \underline{3.23} & \underline{3.9} & \textbf{0.83} & \underline{1.39} & \underline{1.82} & \underline{2.25} \\
        Our & \underline{27.16} & \textbf{45.97} & \textbf{64.76} & \textbf{87.36} & \textbf{1.54} & \textbf{2.25} & \textbf{2.65} & \textbf{3.01} & \underline{0.85} & \textbf{1.23} & \textbf{1.41} & \textbf{1.65} \\
        \bottomrule
    \end{tabular*}
    \label{tab:2}
\end{table*}

In terms of intensity prediction, our model achieved a 53\% to 71.8\% improvement over the CMO forecasts, and notably, this advantage remained consistent across all forecast horizons. Furthermore, our method demonstrated competitive results compared to other deep learning-based models. Although our 6-hour trajectory forecast error was slightly higher than that of the recent MGTCF model, our approach consistently achieved state-of-the-art performance at longer forecast intervals.

It is worth noting that several advanced deep learning models that have shown strong performance in pedestrian and vehicle trajectory prediction did not achieve comparable success in TC forecasting. This discrepancy is largely attributed to the fundamentally different nature of the tasks: pedestrian and vehicle trajectory prediction primarily focuses on road conditions and inter-agent social interactions, whereas TC prediction is governed by far more complex physical processes. The strength of our model lies in its ability to process significantly larger and more diverse datasets than prior approaches. Our findings highlight a strong correlation between forecasting accuracy and the availability of relevant meteorological data—more data enables the model to extract richer, more informative features, thereby improving prediction performance.

Building upon our initial evaluation, we extended the TC forecasting experiments to include data from 2020 to 2024 and conducted additional comparisons with MMSTN and MGTCF. As shown in Table \ref{tab:2}, our method continued to demonstrate strong performance, consistent with the results observed in the 2017–2019 period. We also observed that extreme values and intensities, which are becoming more frequent under global warming, tend to be underestimated when using limited training datasets. As climate change intensifies, it becomes increasingly critical to continuously update models with the most recent data to ensure accurate predictions. Finally, Fig. \ref{fig:5} presents case studies of TC forecasts from 2024, further validating the superior performance of our proposed model.

\subsection{Comparison With Global Operational Forecasting Methods}
In addition to deep learning-based approaches, we further compared our model against NWP systems, including forecasts from official meteorological agencies across multiple countries, as well as both global and regional models from 2020 to 2022. As summarized in Table \ref{tab:3}, our method achieved state-of-the-art performance in intensity forecasting and demonstrated competitive accuracy in 12-hour trajectory predictions. Although our trajectory forecasts did not surpass the best-performing NWP models, our approach outperformed several NWP systems while operating with only a single GPU. This result highlights the rapid-response capability of our model in ultra-short-term forecasting scenarios and underscores its robustness in resource-constrained environments. The ability to deliver accurate predictions with limited computational resources indicates the model's scalability and practical applicability for real-world deployment.

\begin{table*}[!htb]
    \centering
    \caption{Comparison with Global Operational Models}
    \begin{tabular*}{\textwidth}{@{\extracolsep{\fill}}c*{12}{c}@{}}
        \toprule
        & \multicolumn{4}{c}{2020} & \multicolumn{4}{c}{2021} & \multicolumn{4}{c}{2022} \\
        \cmidrule(lr){2-5} \cmidrule(lr){6-9} \cmidrule(lr){10-13}
        & \multicolumn{2}{c}{Distance (km)} & \multicolumn{2}{c}{Wind (m/s)} & 
        \multicolumn{2}{c}{Distance (km)} & \multicolumn{2}{c}{Wind (m/s)} &
        \multicolumn{2}{c}{Distance (km)} & \multicolumn{2}{c}{Wind (m/s)} \\
        \cmidrule(lr){2-3} \cmidrule(lr){4-5} \cmidrule(lr){6-7} \cmidrule(lr){8-9} \cmidrule(lr){10-11} \cmidrule(lr){12-13}
        Methods & 12h & 24h & 12h & 24h & 12h & 24h & 12h & 24h & 12h & 24h & 12h & 24h \\
        \midrule
        CMA & - & 73.5 & - & 4.5 & 56.7 & 88.3 & - & 4.4 & 51.9 & 76.5 & - & 4.5 \\
        JMA & - & 74.2 & - & 4.5 & 51.3 & 85.3 & - & 4.4 & 48.4 & 72.7 & - & 5.5 \\
        JTWC & - & 74.9 & - & 4.9 & 56.6 & 89.3 & - & 4.8 & 55.6 & 78.2 & - & 5.4 \\
        KMA & - & 88.3 & - & 4.7 & 61 & 97.9 & - & 4.9 & 59.8 & 83.8 & - & 5.1 \\
        HKO & - & 70.6 & - & 4.7 & - & 83.2 & - & 4.9 & - & 79.3 & - & 5.2 \\
        \midrule
        CMA-GFS & - & - & - & - & 74.5 & 115.5 & - & 10.0 & 66.6 & 88.3 & - & 10.6 \\
        NCEP-GFS & 47.4 & 66.8 & 4.1 & 4.7 & 49.7 & 74.4 & - & 5.4 & 48.7 & 67.4 & - & 6.7 \\
        ECMWF-IFS & 43.7 & \textbf{58.0} & 6.5 & 6.7 & \textbf{42.2} & \textbf{61.1} & - & 7.5 & 46.3 & 63.3 & - & 11.0 \\
        UKMD-MetUM & 46.1 & 65.5 & 5.3 & 6.1 & 49 & 73.5 & - & 8.1 & 49.2 & 68.6 & - & 7.8 \\
        JMA-GSM & 51.4 & 74.1 & 4.1 & 6.0 & 55.1 & 95.2 & - & 5.5 & 49.1 & 70.5 & - & 6.5 \\
        \midrule
        STI-TEDAPS & 46.2 & 60.8 & 5.6 & 5.8 & 58.6 & 82.1 & - & 7.6 & 55.9 & 78.7 & - & 8.4 \\
        CMA-TRAMS & 46.1 & 61.3 & 4.0 & 5.2 & 47.6 & 67.2 & - & 5.8 & \textbf{42.4} & \textbf{55.3} & - & 5.8 \\
        CMA-TYM & 48.7 & 77.3 & 4.4 & 5.2 & 59.3 & 86.7 & - & 4.8 & 53 & 78.9 & - & 6.4 \\
        CMA-TCM & 66.2 & 99.8 & 4.2 & 5.7 & 77.9 & 107.2 & - & 5.4 & - & - & - & - \\
        HWRF & 53.3 & 73.0 & 3.9 & 4.8 & - & - & - & - & 58.2 & 83.7 & - & 6.2 \\
        \midrule
        Our & \textbf{42.04} & 80.86 & \textbf{1.28} & \textbf{1.72} & 49.57 & 88.3 & \textbf{1.17} & \textbf{1.62} & 48.94 & 104.72 & \textbf{1.26} & \textbf{1.56} \\
        \bottomrule
    \end{tabular*}
    \label{tab:3}
\end{table*}

\subsection{Ablation Study}
To comprehensively evaluate the independent contributions and synergistic effects of the AOT-TCs dataset and the internal components of AOT-TCNet, we conducted systematic ablation experiments on the CMA-BST dataset. Table \ref{tab:4} presents the performance variations after introducing the TMA-MoE module and verifies the effectiveness of the heterogeneous atmospheric/oceanic/terrestrial (A/O/T) variables proposed in this study. When the model was trained using only ERA5 inputs, both the track error and intensity error remained at relatively high levels. After incorporating the TMA-MoE module, the forecasting accuracy at all lead times improved significantly. This indicates that explicitly identifying and modeling different typhoon track modes can effectively alleviate the instability caused by mode mixing and enhance the model’s adaptability to different stages of dynamical evolution. We further analyzed the contributions of the ocean- and land-related variables in the AOT-TCs dataset. When only atmospheric variables were used, the model achieved reasonable performance. After oceanic information was introduced, most evaluation metrics improved substantially, reflecting the critical role of upper-ocean thermodynamic conditions in regulating TC energy uptake and translation speed. More importantly, after further incorporating land-related variables on top of the A+O configuration, the model performance improved consistently across all forecast lead times. This demonstrates that topographic effects, including channeling effects, blocking effects, and deflection effects, are indispensable for accurately characterizing anomalous typhoon behavior. When TMA-MoE and A/O/T were enabled simultaneously, the model achieved the best overall performance. These results indicate that TMA-MoE enhances the model’s ability to adapt to the multimodal evolutionary patterns of TCs, whereas AOT-TCs strengthens physical consistency and mitigates the dynamical biases introduced by purely data-driven learning.

In summary, the heterogeneous A/O/T physical variables provide critical cross-sphere coupling information, enabling the model to better capture the three-dimensional dynamical and thermodynamical evolution of real TCs. The joint integration of AOT-TCs and TMA-MoE allows AOT-TCNet to achieve significant improvements over existing state-of-the-art baselines in forecasting track, central pressure, and maximum wind speed.

\begin{table*}[!htb]
    \centering
    \caption{Comparative Results from the Ablation Study}
    \begin{tabular*}{\textwidth}{@{\extracolsep{\fill}}cccc*{12}{c}@{}}
        \toprule
        \multicolumn{4}{c}{Components} & \multicolumn{4}{c}{Distance (km)} & \multicolumn{4}{c}{Pressure (hpa)} & \multicolumn{4}{c}{Wind (m/s)} \\
        \cmidrule(lr){1-4} \cmidrule(lr){5-8} \cmidrule(lr){9-12} \cmidrule(lr){13-16}
        A & M & O & T & 6h & 12h & 18h & 24h & 6h & 12h & 18h & 24h & 6h & 12h & 18h & 24h \\
        \midrule
        \  & \  & \  & \ & 27.99 & 60.76 & 97.85 & 141.73 & 1.94 & 3.23 & 4.34 & 5.18 & 0.95 & 1.58 & 2.11 & 2.58 \\
        \checkmark & \  & \  & \ & 25.98 & 49.67 & 74.42 & 105.72 & 1.31 & 2.01 & 2.73 & 3.23 & 0.73 & 1.18 & 1.54 & 1.85 \\
        \checkmark & \checkmark & \  & \ & 25.45 & 47.28 & 71.16 & 98.03 & 1.39 & 2.11 & 2.81 & 3.44 & 0.77 & 1.21 & 1.58 & 1.86 \\
        \checkmark & \checkmark  & \checkmark  & \ & 25.08 & 42.63 & 60.32 & 83.95 & 1.21 & 1.87 & 2.27 & 2.7 & 0.71 & 1.06 & 1.27 & 1.46 \\
        \checkmark & \checkmark & \  & \checkmark & 26.94 & 47.34 & 70.41 & 100.76 & 1.24 & 1.78 & 2.22 & 2.61 & 0.69 & 1.04 & 1.32 & 1.52 \\
        \checkmark & \checkmark & \checkmark & \checkmark & 24.03 & 41.98 & 59.58 & 86.05 & 1.22 & 1.74 & 2.19 & 2.58 & 0.71 & 1.03 & 1.23 & 1.41 \\
        \bottomrule
    \end{tabular*}
    \label{tab:4}
\end{table*}

\begin{table}[!htb]
    \centering
    \caption{Comparison of different methods for abnormally deflected TC prediction}
    \begin{tabular}{c*{6}{c}}
        \toprule
        & \multicolumn{2}{c}{Distance (km)} & \multicolumn{2}{c}{Pressure (hpa)} & \multicolumn{2}{c}{Wind (m/s)} \\
        \cmidrule(lr){2-3} \cmidrule(lr){4-5} \cmidrule(lr){6-7}
        Method & 12h & 24h & 12h & 24h & 12h & 24h \\
        \midrule
        MMSTN & 73.61 & 173.72 & 3.04 & 5.26 & 1.52 & 2.64 \\
        MGTCF & 49.72 & 107.80 & 2.27 & 4.05 & 1.23 & 2.06 \\
        Our & \textbf{47.39} & \textbf{93.29} & \textbf{2.02} & \textbf{3.04} & \textbf{1.02} & \textbf{1.56} \\
        \bottomrule
    \end{tabular}
    \label{tab:6}
\end{table}
\begin{figure*}[!htb]
    \centering
    \includegraphics[width=0.7\textwidth]{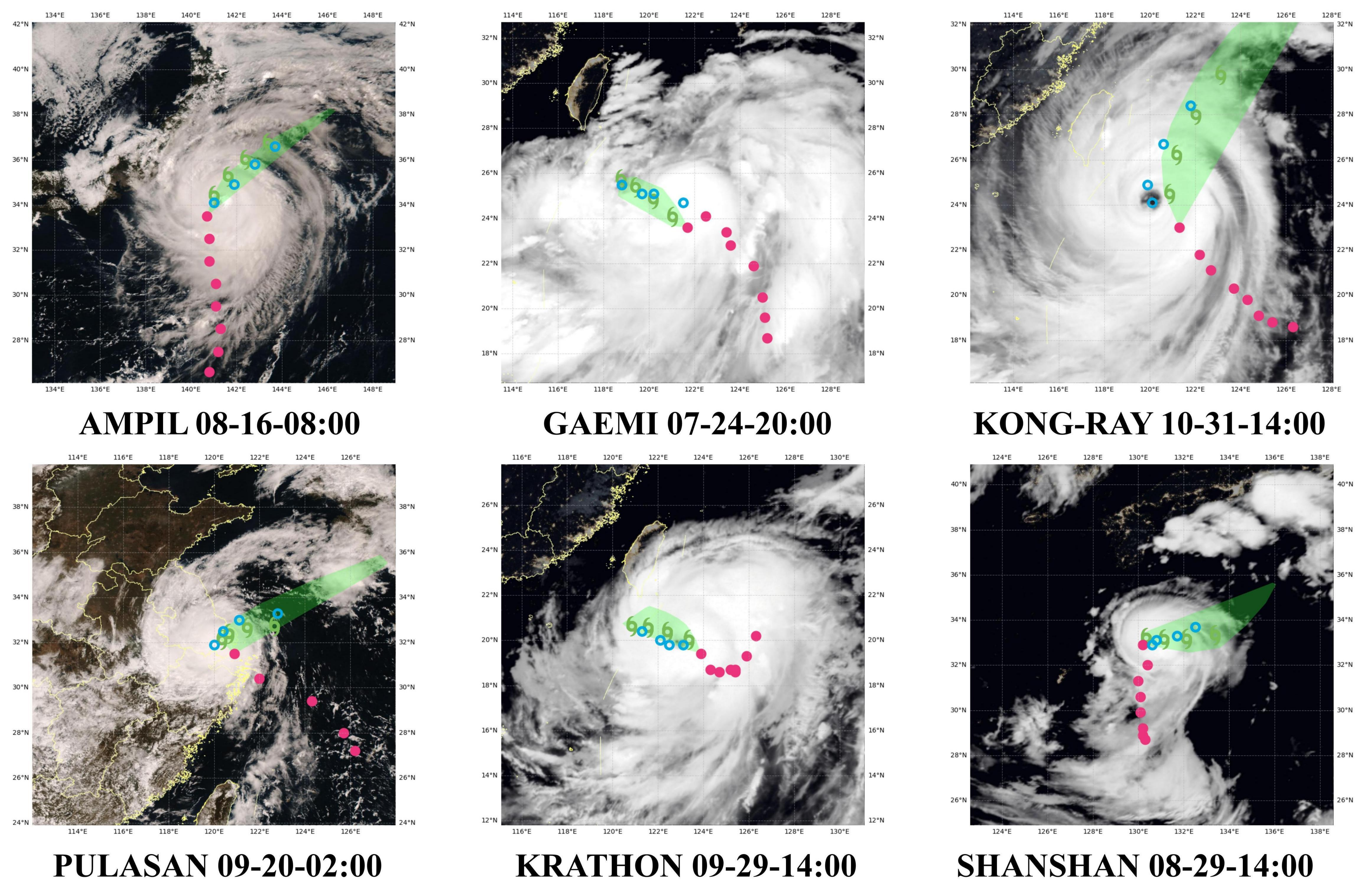}
    \caption{Ensemble forecasting results for abnormally deflected TCs in 2024. Historical trajectories are shown in red, actual future tracks in blue, predicted paths in green, and shaded regions denote the projected trajectory trends.} 
    \label{fig:6} 
\end{figure*}
\subsection{Ensemble Forecasting of Abnormally Deflected TCs}
Abnormally deflected TCs refer to those that exhibit abrupt trajectory changes within a short time span due to the combined influence of multiple meteorological and environmental factors, including blocking highs, westerly troughs, the terrain of Taiwan Island, and binary TC interactions. In this study, we define an abnormal deflection event as a rightward turning angle exceeding 45° or a leftward angle exceeding 30° within a 12-hour period. These TCs are characterized by limited sample availability and rapid path shifts, posing significant challenges for prediction. Existing forecasting models often struggle with these cases due to over-smoothed outputs and the absence of physically grounded constraints, making it difficult to capture such sharp turning behaviors. To address this, we leveraged a multi-generator framework combined with surface elevation information to enhance the model's capability for abnormal scenario simulation and causal reasoning. Our approach successfully predicted the behavior of abnormally deflected TCs. We identified and reported all such cases in the test set individually, and the results, presented in Table \ref{tab:6}, show that our method consistently achieved the best performance. These findings underscore the critical role of physical variables and generative modeling in predicting anomalous TC deflections.
Furthermore, we conducted a detailed case analysis of the abnormally deflected TCs in 2024, accompanied by ensemble forecast visualizations. As illustrated in Fig. \ref{fig:6}, AOT-TCNet effectively captured both the sudden deflection and the overall deflection trend. Although some positional errors remain, the results clearly demonstrate the model’s robust capability for ensemble forecasting of abnormally deflected TCs.

\subsection{Case Analysis}
\label{sec:case analysis}
We focus on analyzing our model’s prediction of super TC YAGI. YAGI formed over the Northwestern Pacific east of the Philippines on September 1, 2024, made landfall in northeastern Philippines on September 2, and later entered the South China Sea. It made another landfall along the coast of Hainan from the afternoon to the night on September 6. During its trajectory, the TC reached a maximum sustained wind intensity of Category 17, resulted in 16 fatalities, and caused nearly 80 billion CNY in direct economic losses.
\begin{figure*}[!htb]
    \centering
    \begin{subfigure}[b]{0.24\textwidth}
        \includegraphics[width=\textwidth]{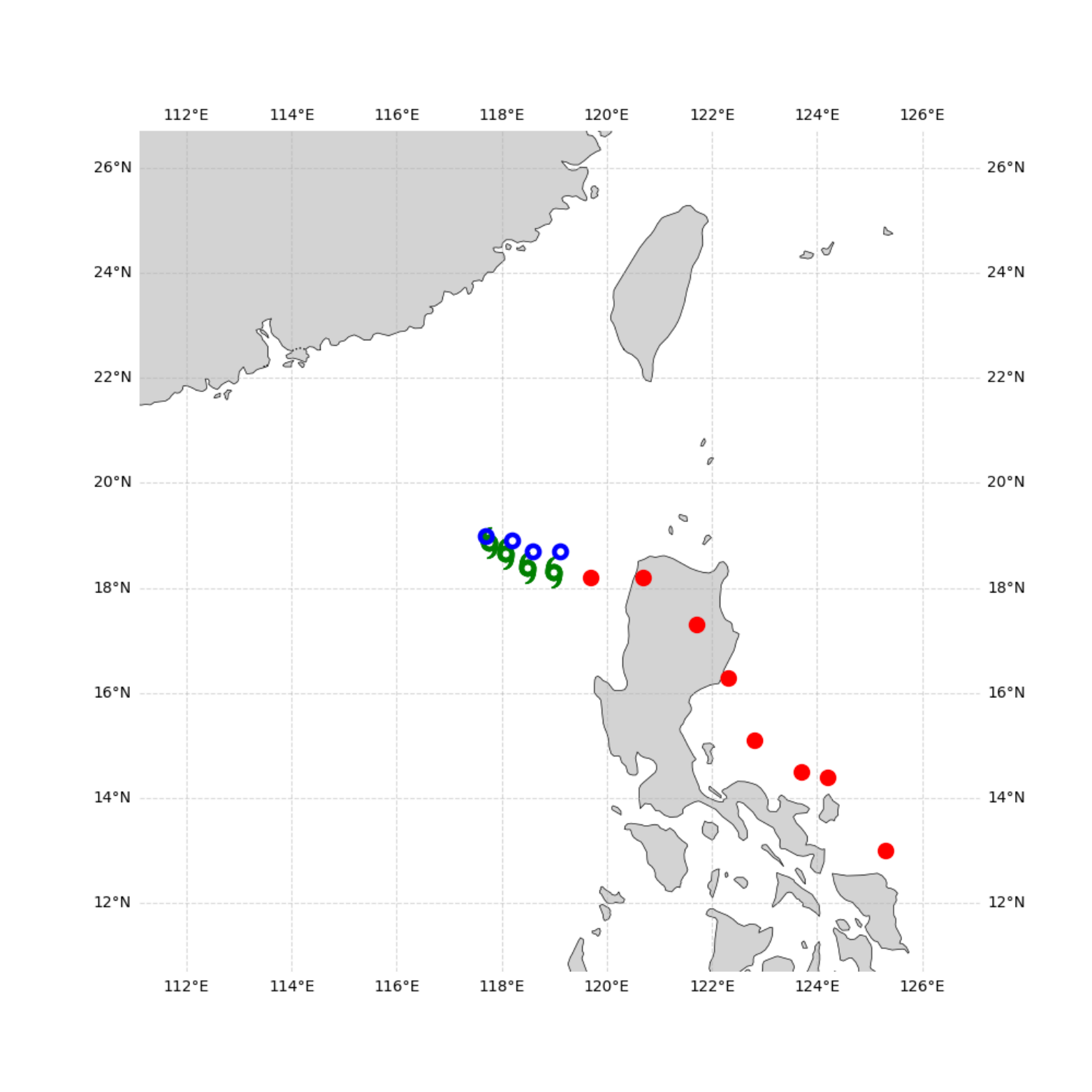}
        \caption{YAGI 09-03-00:00}
        \label{fig:7-1}
    \end{subfigure}
    \hfill
    \begin{subfigure}[b]{0.24\textwidth}
        \includegraphics[width=\textwidth]{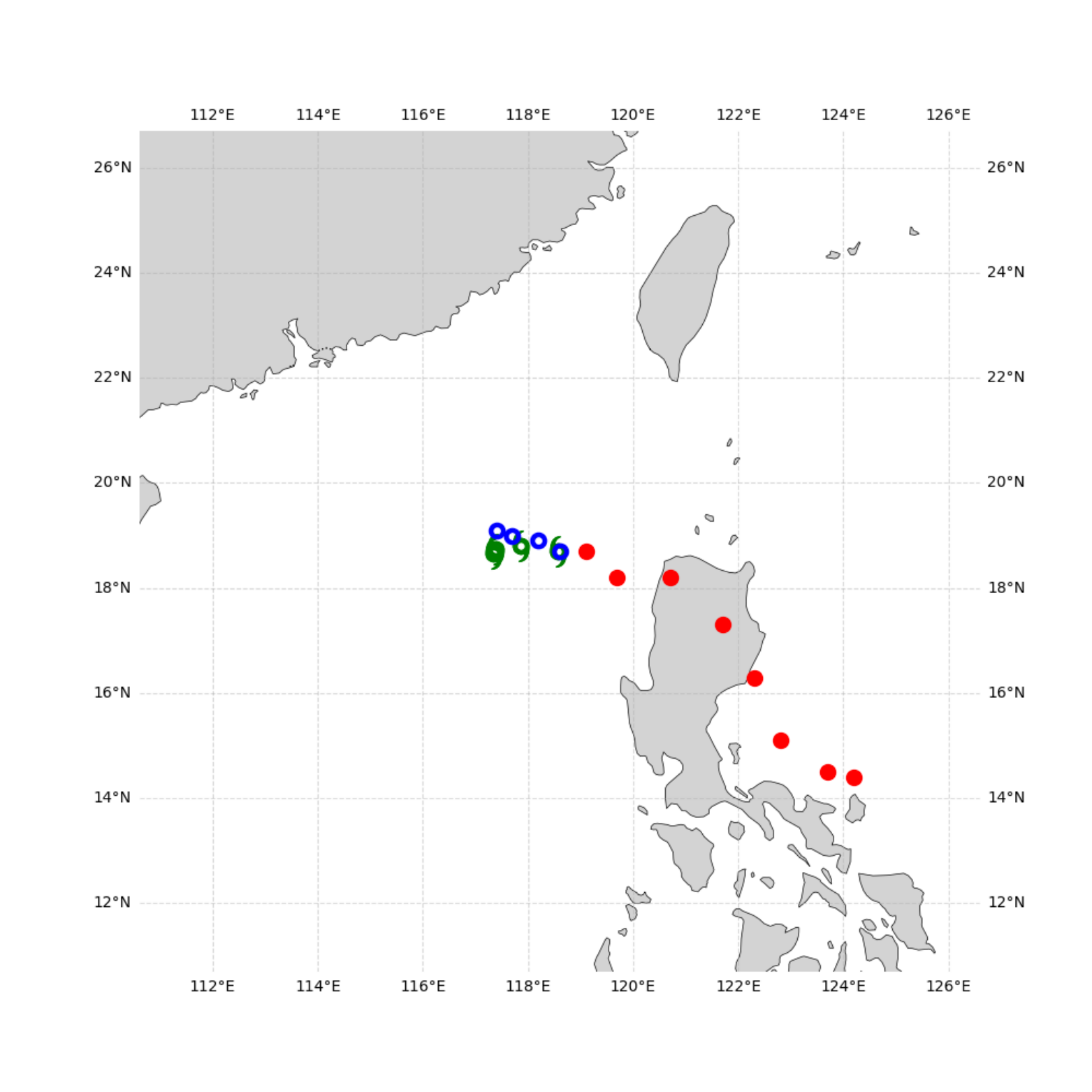}
        \caption{YAGI 09-03-06:00}
        \label{fig:7-2}
    \end{subfigure}
    \hfill
    \begin{subfigure}[b]{0.24\textwidth}
        \includegraphics[width=\textwidth]{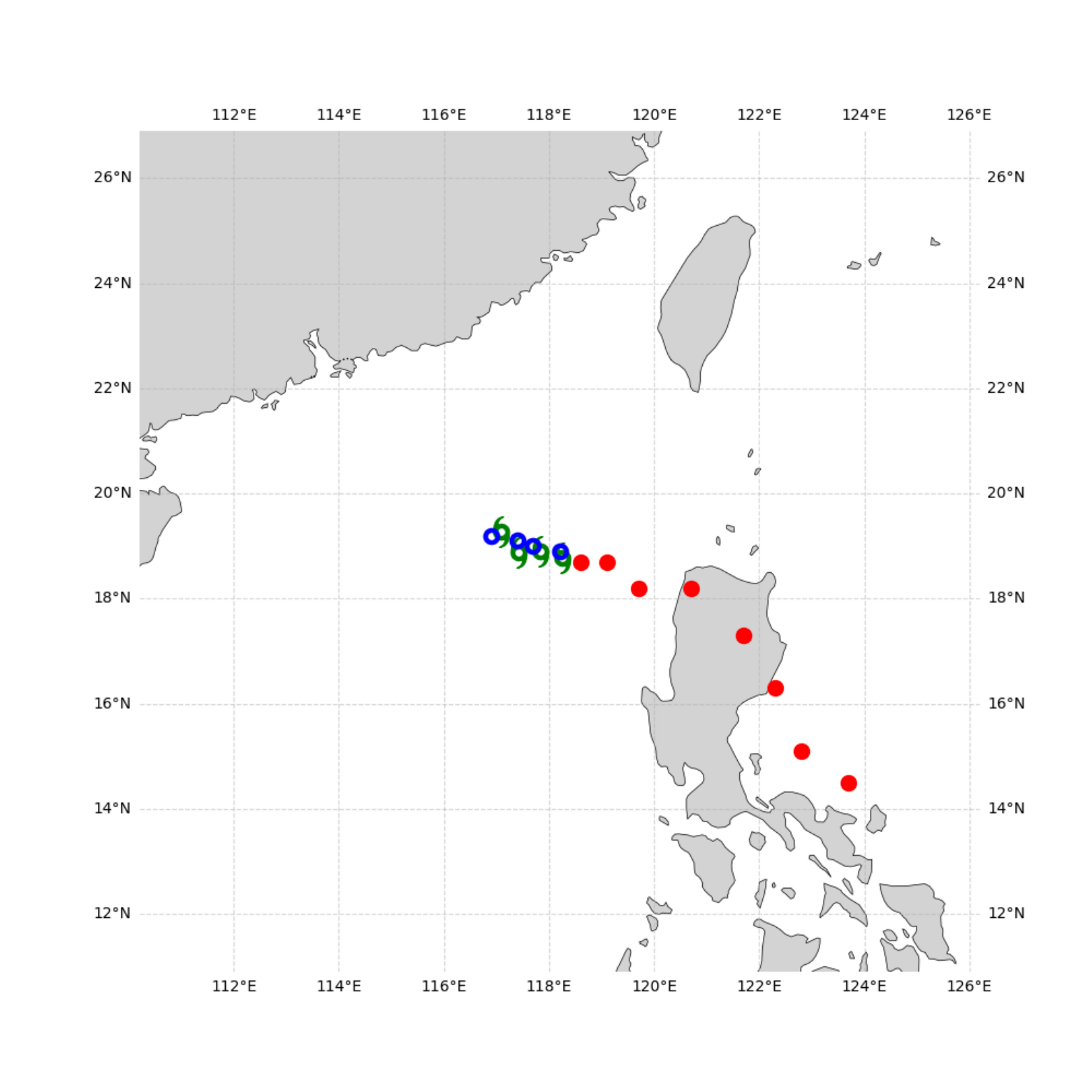}
        \caption{YAGI 09-03-12:00}
        \label{fig:7-3}
    \end{subfigure}
    \hfill
    \begin{subfigure}[b]{0.24\textwidth}
        \includegraphics[width=\textwidth]{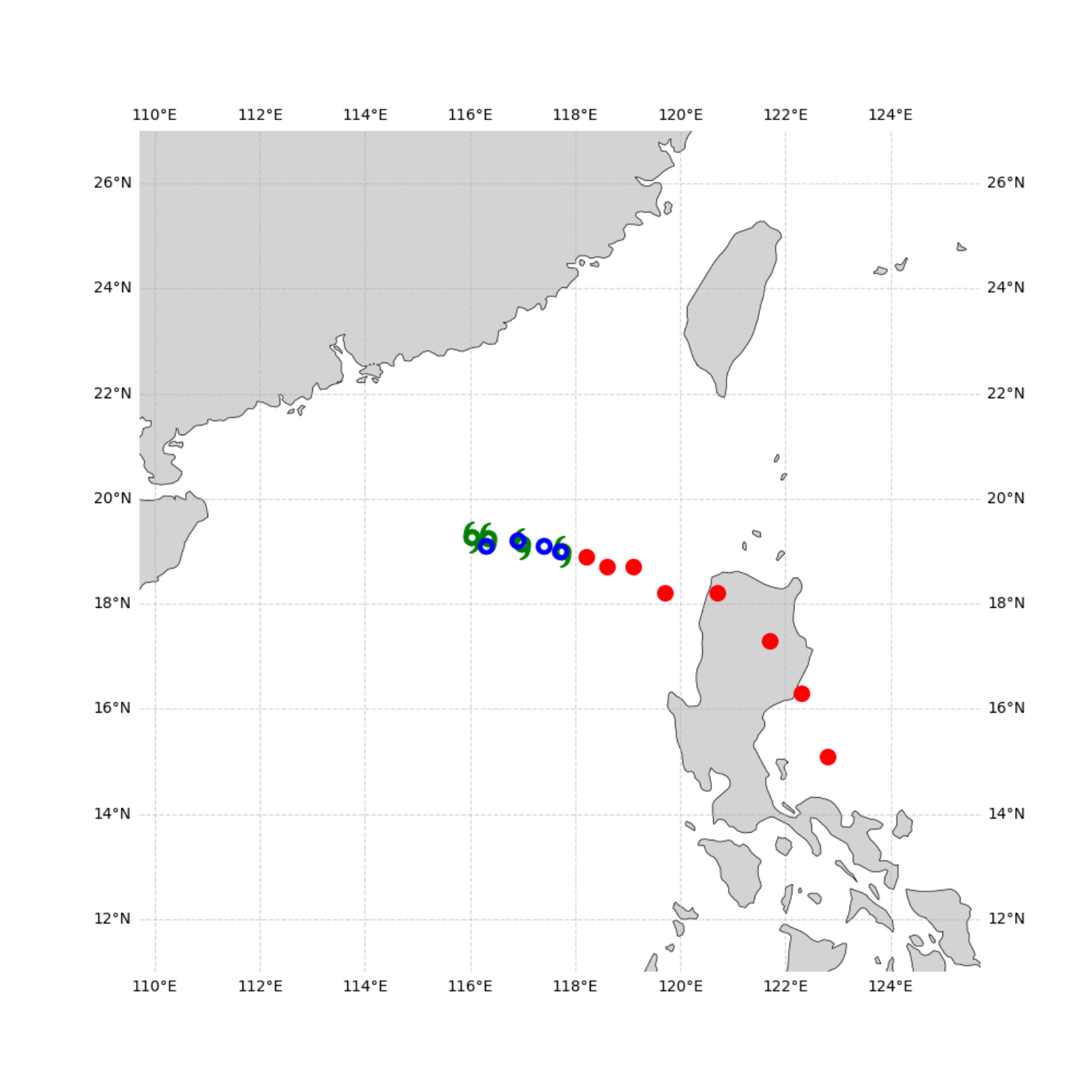}
        \caption{YAGI 09-03-18:00}
        \label{fig:7-4}
    \end{subfigure}
    
    \begin{subfigure}[b]{0.24\textwidth}
        \includegraphics[width=\textwidth]{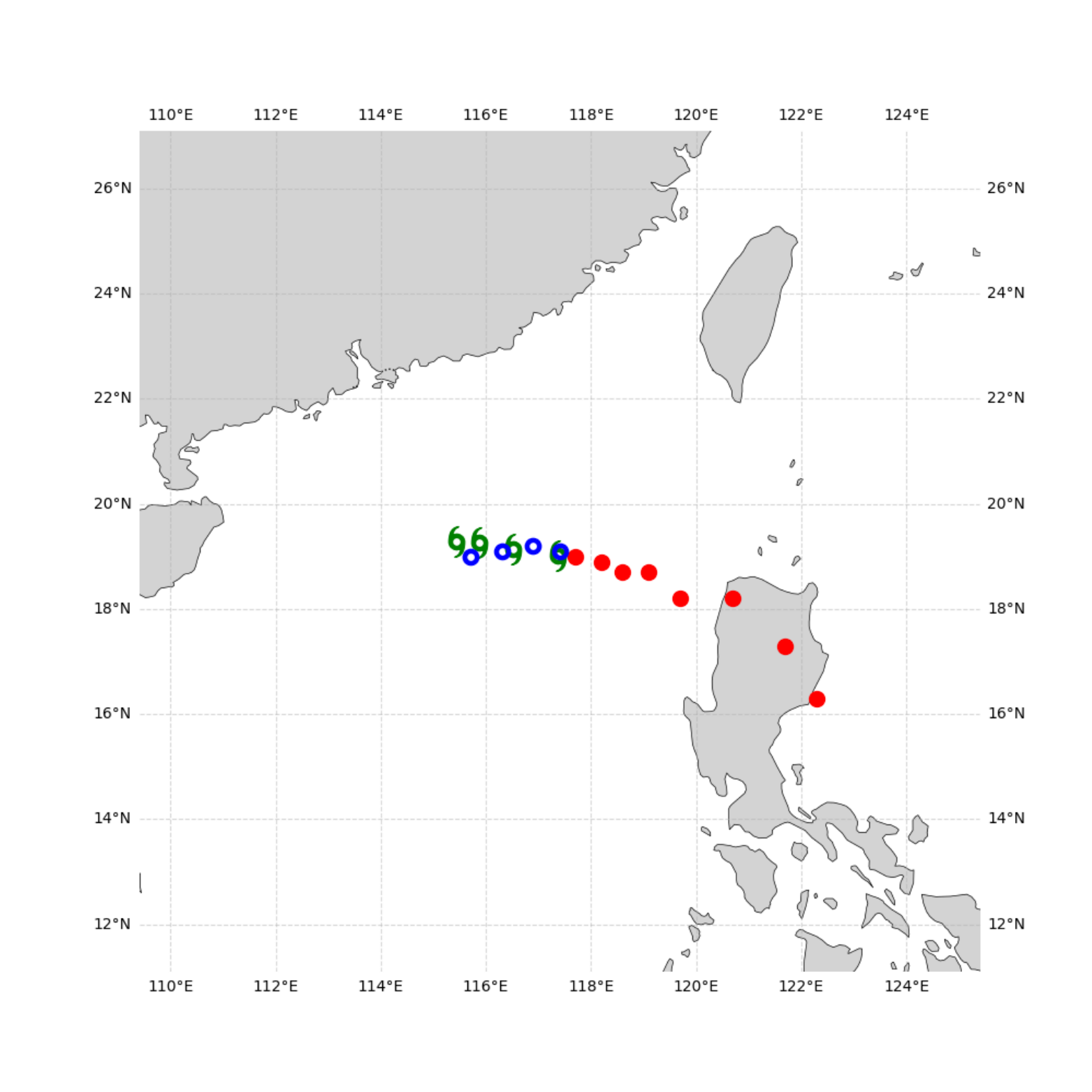}
        \caption{YAGI 09-04-00:00}
        \label{fig:7-5}
    \end{subfigure}
    \hfill
    \begin{subfigure}[b]{0.24\textwidth}
        \includegraphics[width=\textwidth]{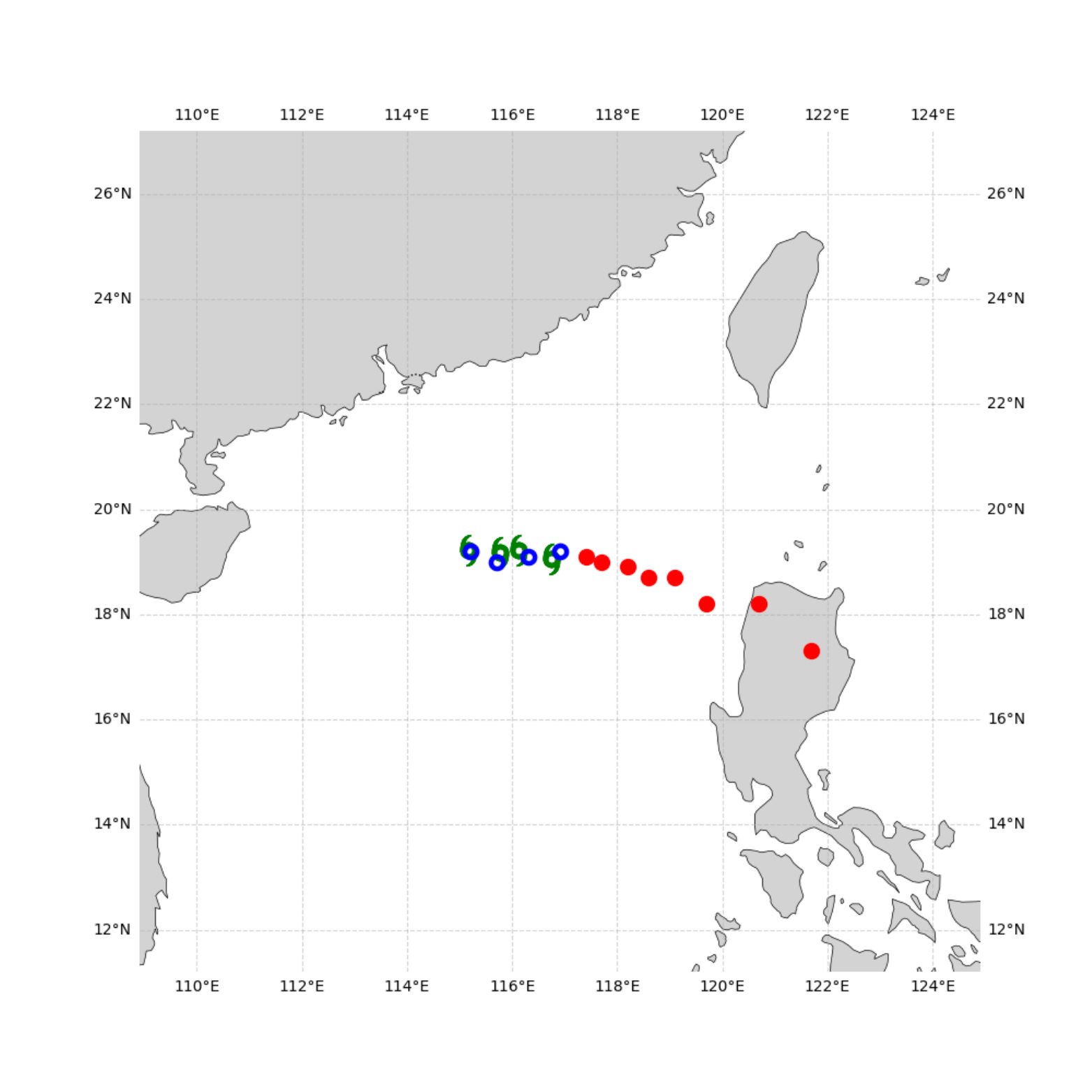}
        \caption{YAGI 09-04-06:00}
        \label{fig:7-6}
    \end{subfigure}
    \hfill
    \begin{subfigure}[b]{0.24\textwidth}
        \includegraphics[width=\textwidth]{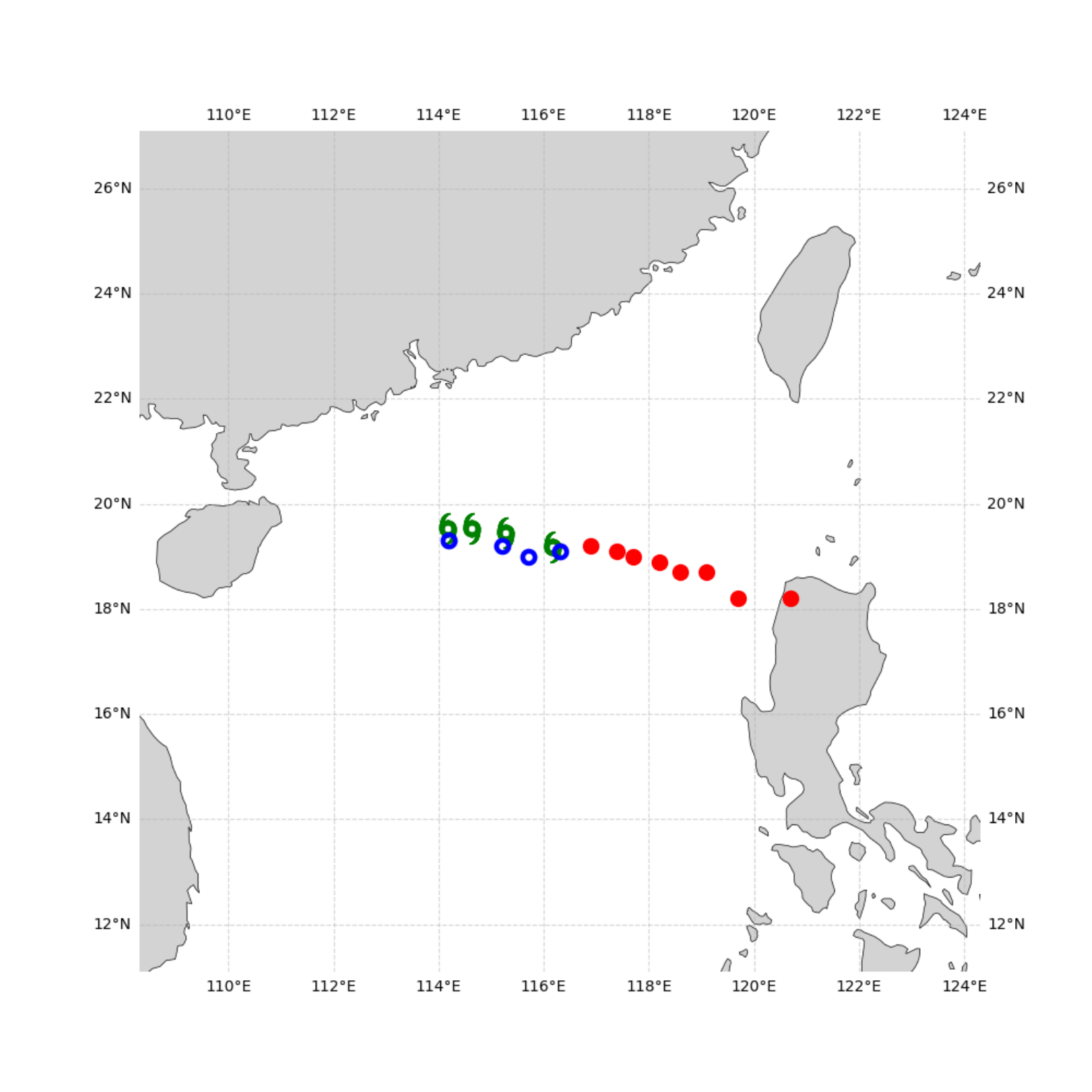}
        \caption{YAGI 09-04-12:00}
        \label{fig:7-7}
    \end{subfigure}
    \hfill
    \begin{subfigure}[b]{0.24\textwidth}
        \includegraphics[width=\textwidth]{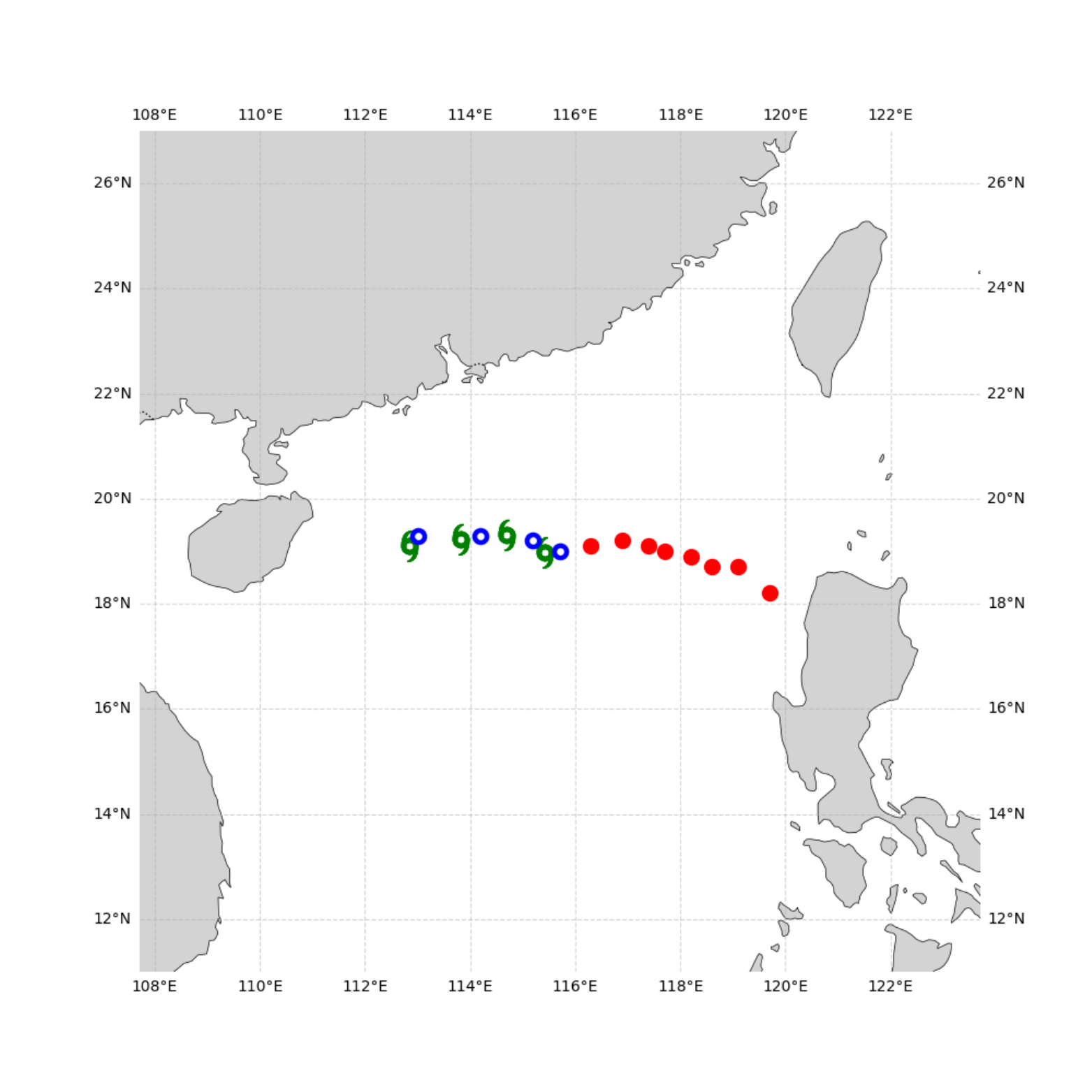}
        \caption{YAGI 09-04-18:00}
        \label{fig:7-8}
    \end{subfigure}
    
    \begin{subfigure}[b]{0.24\textwidth}
        \includegraphics[width=\textwidth]{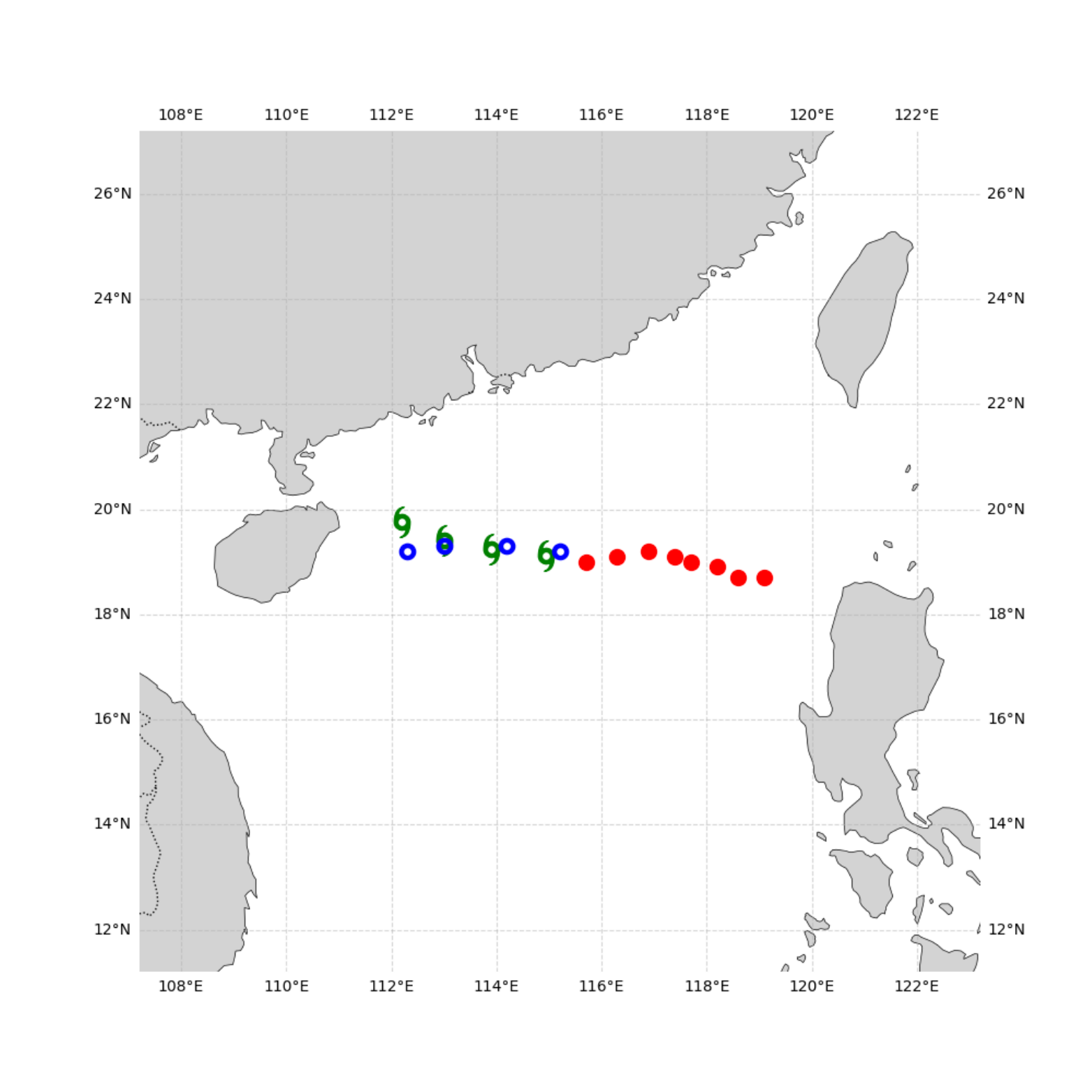}
        \caption{YAGI 09-05-00:00}
        \label{fig:7-9}
    \end{subfigure}
    \hfill
    \begin{subfigure}[b]{0.24\textwidth}
        \includegraphics[width=\textwidth]{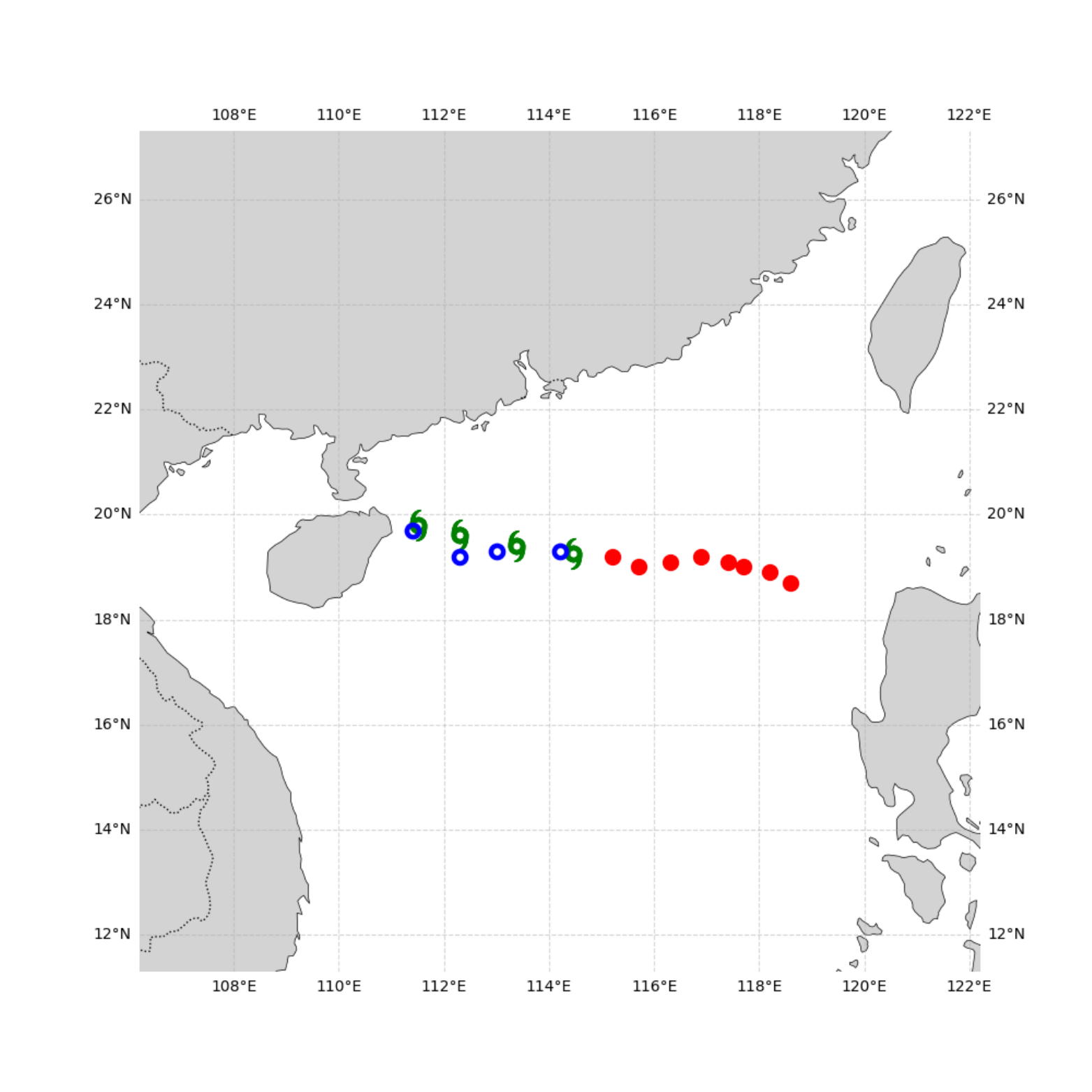}
        \caption{YAGI 09-05-06:00}
        \label{fig:7-10}
    \end{subfigure}
    \hfill
    \begin{subfigure}[b]{0.24\textwidth}
        \includegraphics[width=\textwidth]{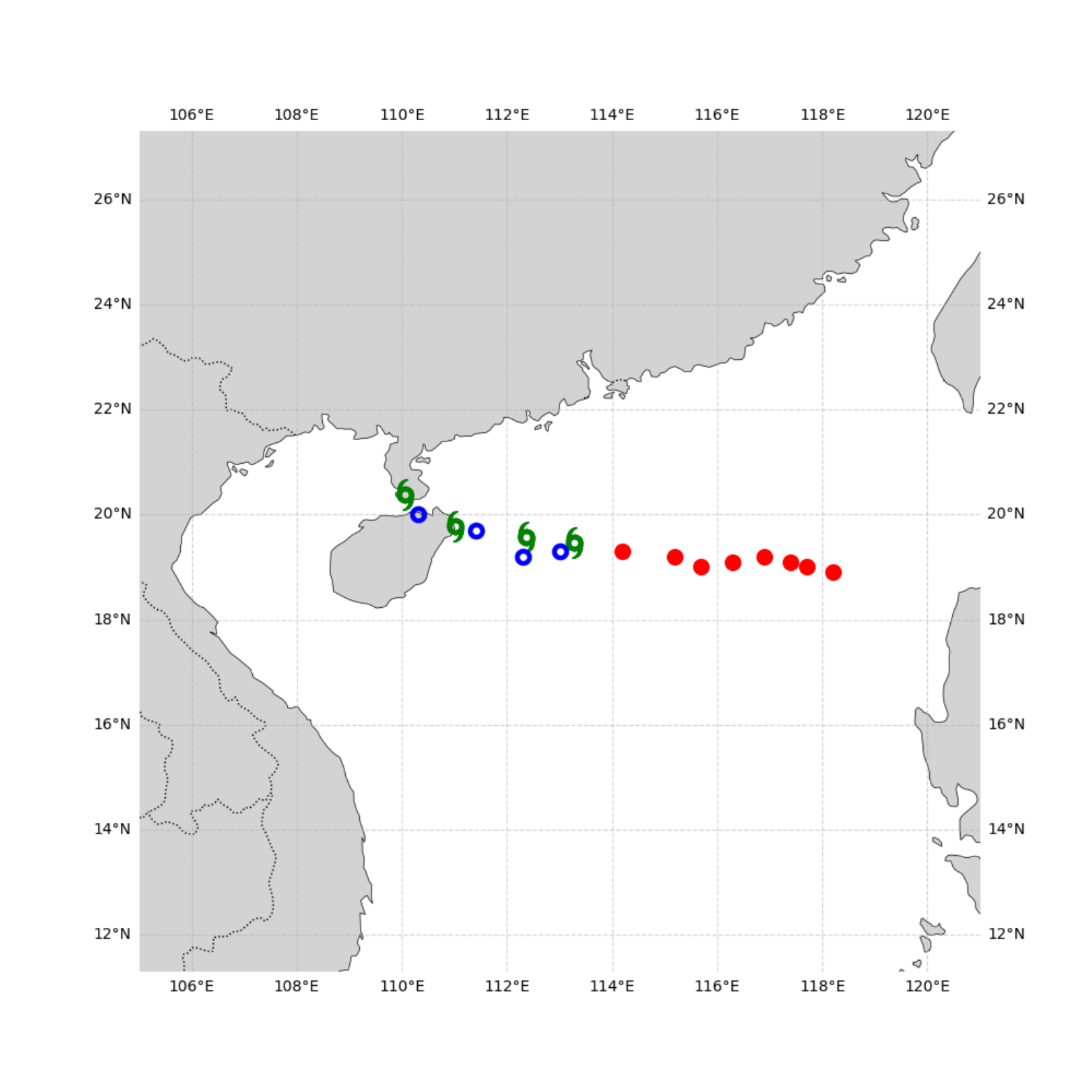}
        \caption{YAGI 09-05-12:00}
        \label{fig:7-11}
    \end{subfigure}
    \hfill
    \begin{subfigure}[b]{0.24\textwidth}
        \includegraphics[width=\textwidth]{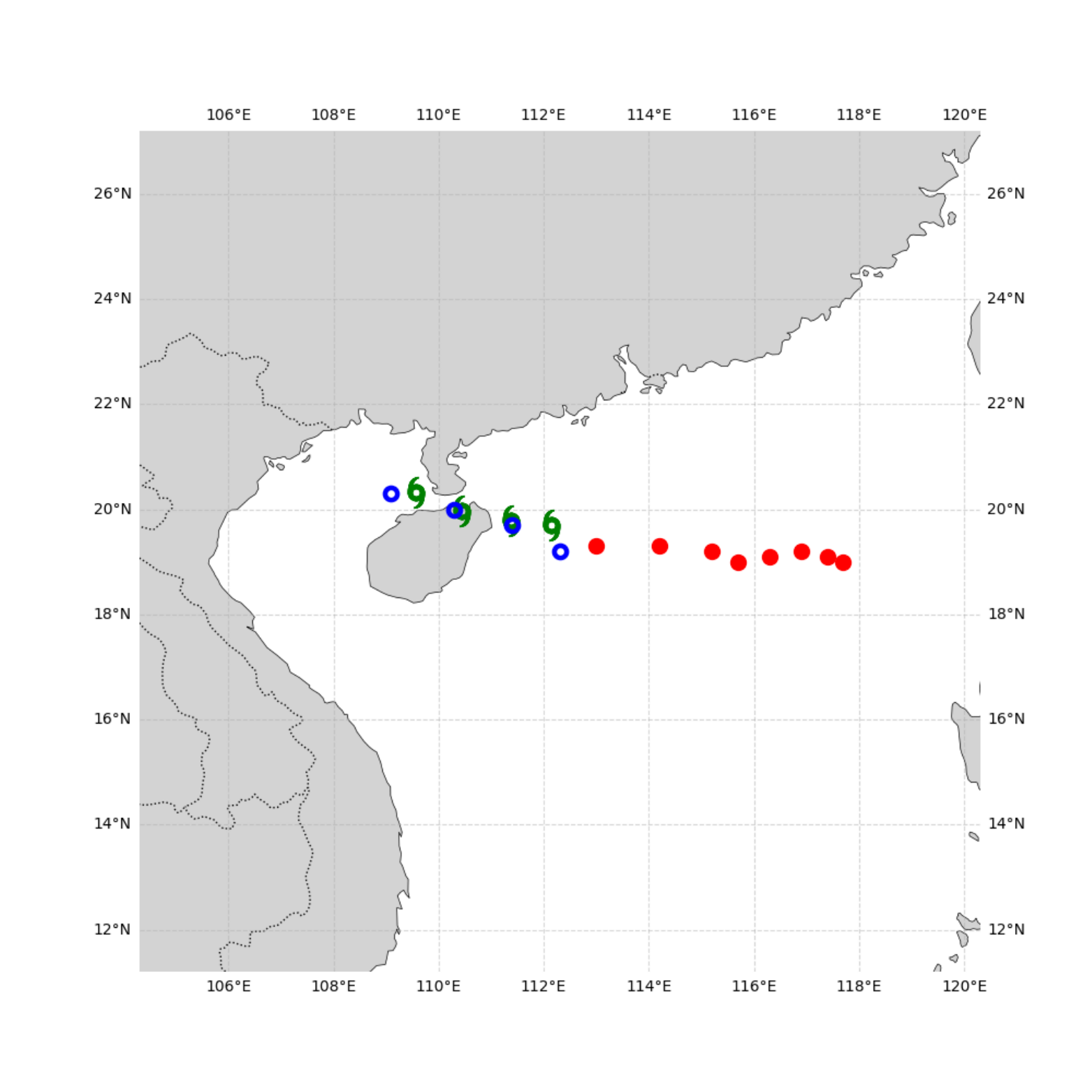}
        \caption{YAGI 09-05-18:00}
        \label{fig:7-12}
    \end{subfigure}
    \hfill
    \begin{subfigure}[b]{0.24\textwidth}
        \includegraphics[width=\textwidth]{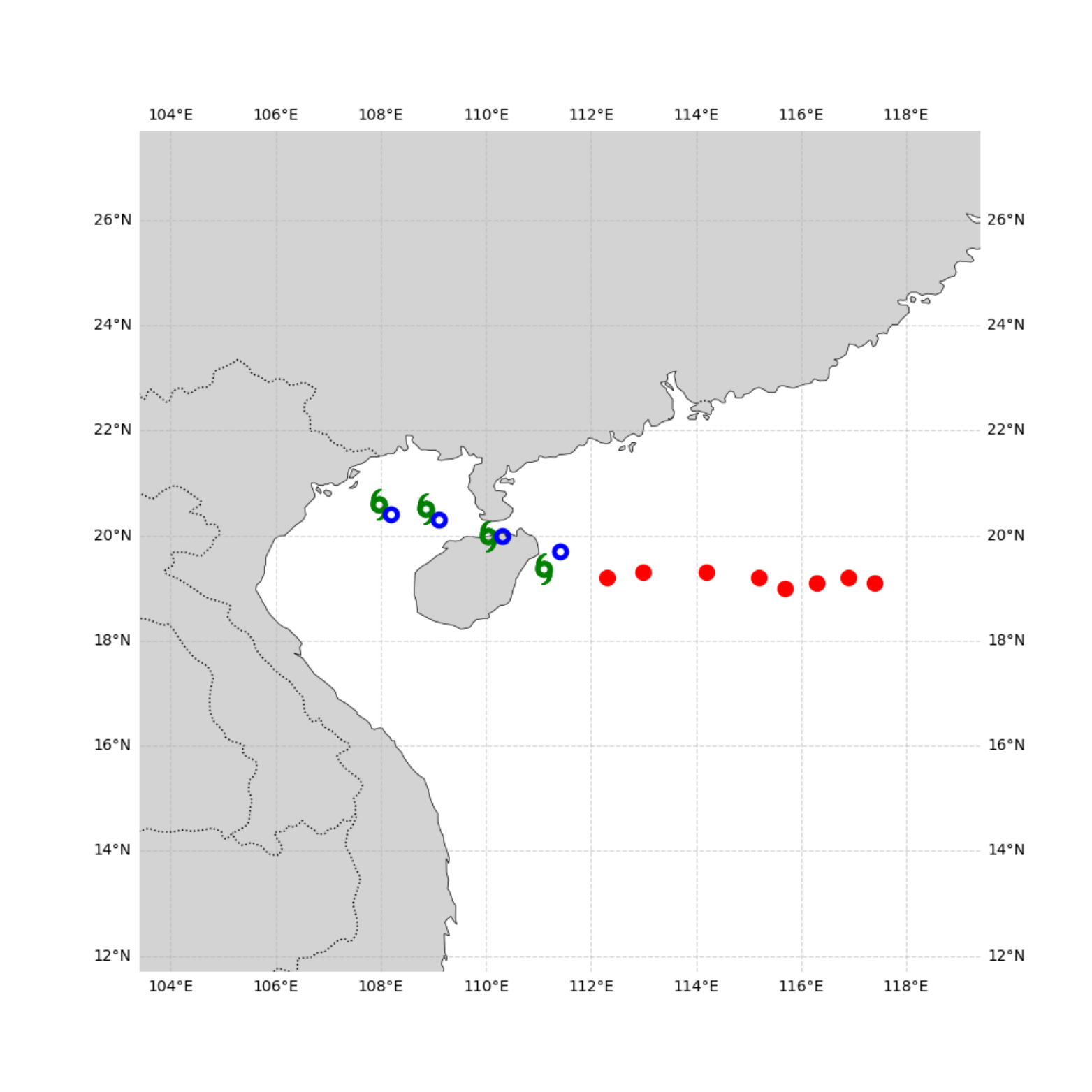}
        \caption{YAGI 09-06-00:00}
        \label{fig:7-13}
    \end{subfigure}
    \hfill
    \begin{subfigure}[b]{0.24\textwidth}
        \includegraphics[width=\textwidth]{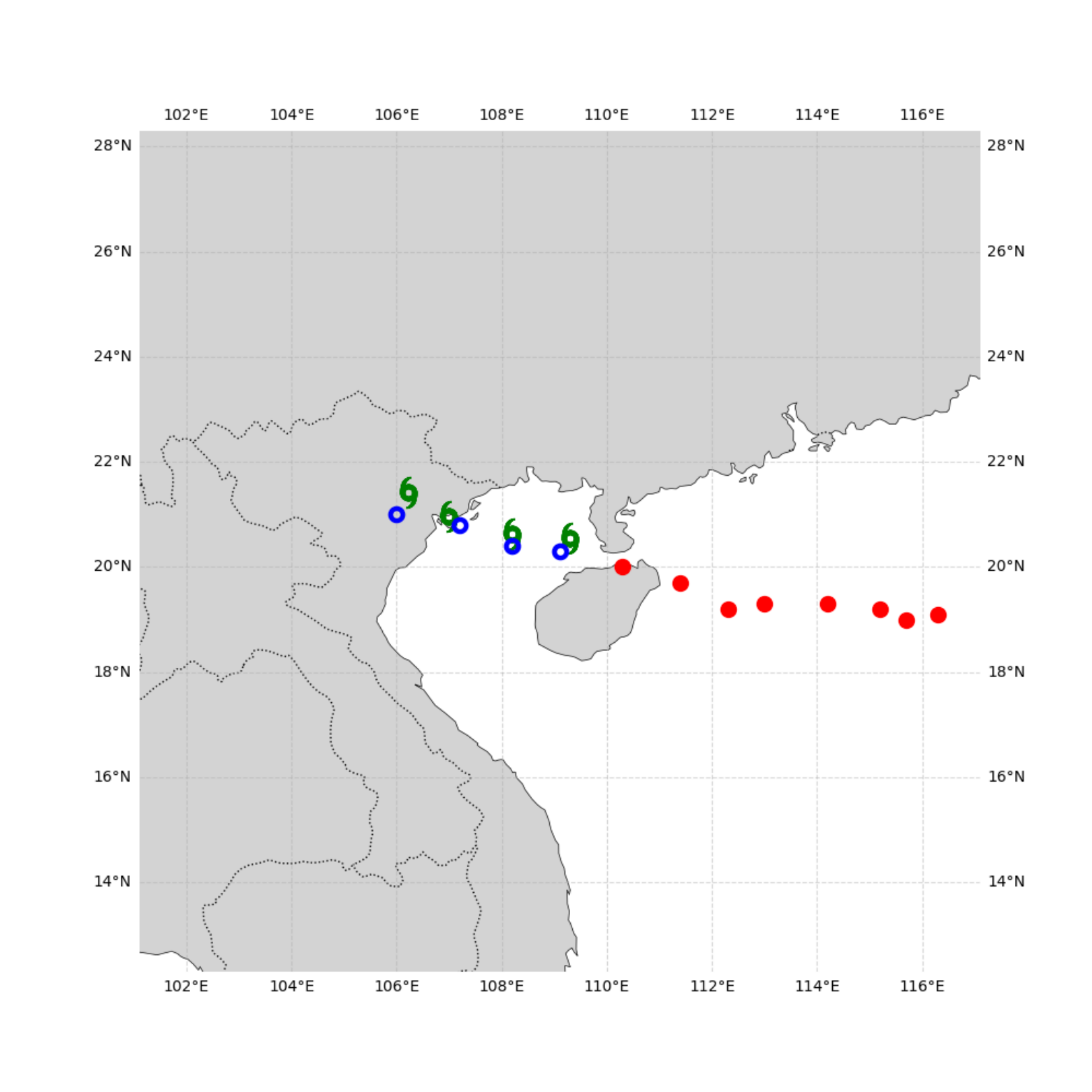}
        \caption{YAGI 09-06-06:00}
        \label{fig:7-14}
    \end{subfigure}
    \hfill
    \begin{subfigure}[b]{0.24\textwidth}
        \includegraphics[width=\textwidth]{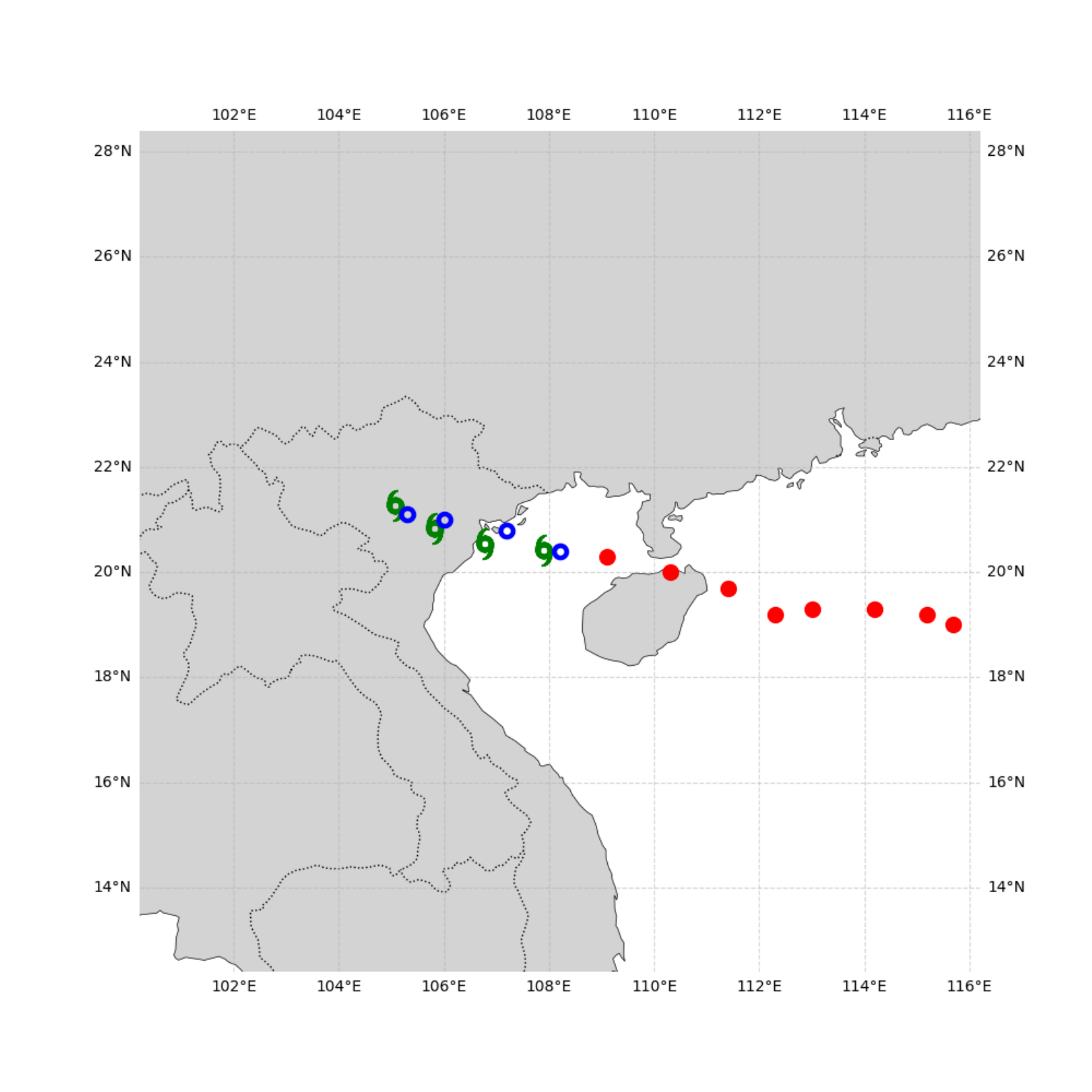}
        \caption{YAGI 09-06-12:00}
        \label{fig:7-15}
    \end{subfigure}
    \hfill
    \begin{subfigure}[b]{0.24\textwidth}
        \includegraphics[width=\textwidth]{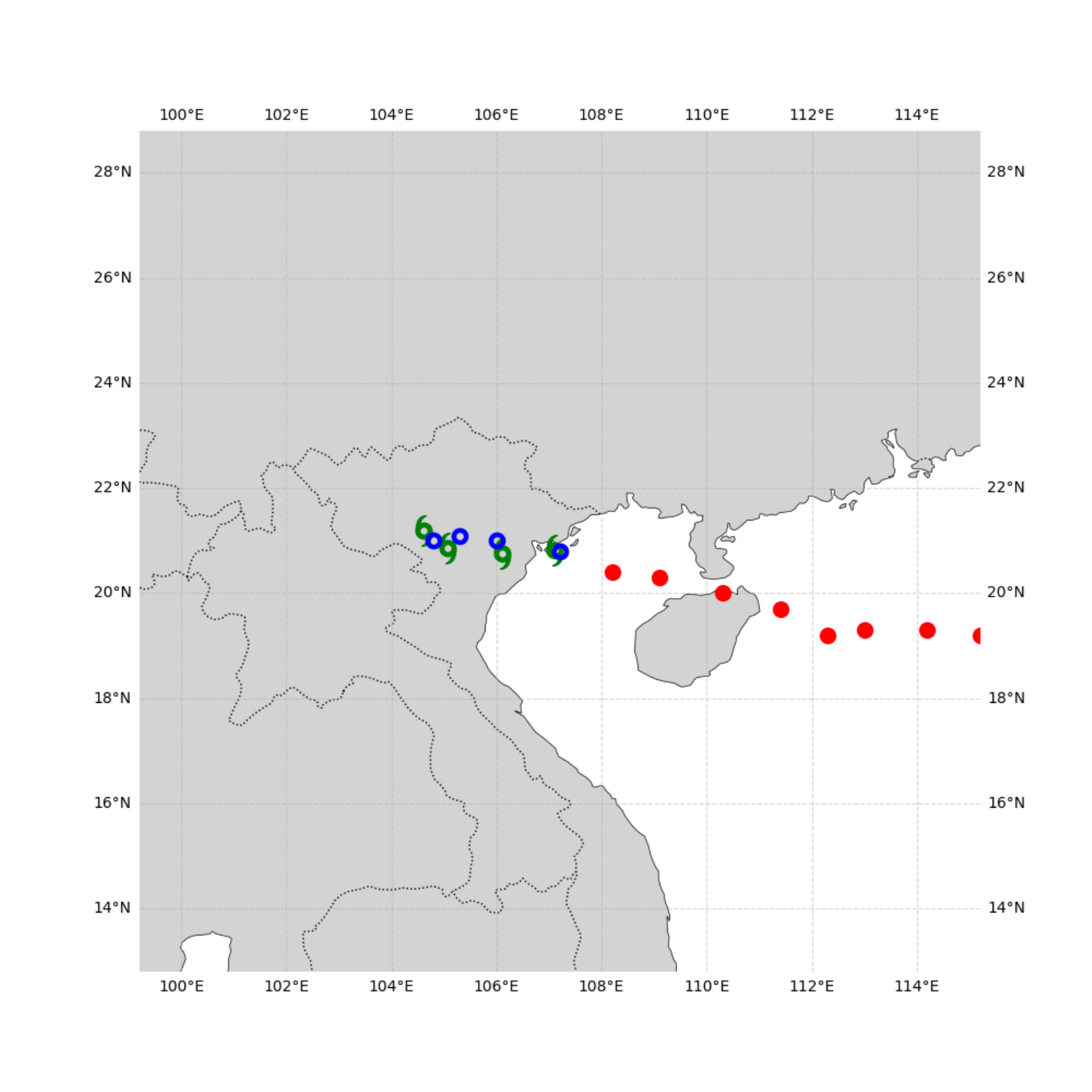}
        \caption{YAGI 09-06-18:00}
        \label{fig:7-16}
    \end{subfigure}
    \hfill
    \caption{Results showcase trajectory predictions using AOT-TCNet for the TC YAGI. The red
dots are input, the green dots are prediction, and the blue dots are GT.}
    \label{fig:7}
\end{figure*}
In the 00Z forecast on August 30, 2024, NCEP-GFS systematically underestimated the intensity of the subtropical high. Due to the high sensitivity of numerical models to initial conditions, even subtle errors can lead to significant deviations in long-term forecasts, which also explains the northward bias in the numerical prediction of TC tracks. NCEP-GFS predicted that YAGI would move north initially, then turn westward and make landfall in northern Fujian. In contrast, ECMWF-IFS HRES forecasted a stronger subtropical high, steering YAGI northwest toward Taiwan before entering Fujian. Our model demonstrated superior performance in predicting both the track and intensity of YAGI, successfully capturing its westward movement,as shown in Figure \ref{fig:7}. In comparison, conventional numerical models exhibited a consistent northward bias in their track forecasts. 

\section{CONCLUSION and Feature Work}
This study proposes the first multimodal TC prediction framework based on atmosphere-ocean-terrain coupled modeling, which improves the prediction accuracy of TC tracks and intensities by fully coupling heterogeneous geophysical variables. In addition, we present a TMA-MoE architecture that enhances forecasting stability through multi-expert forecasting and physical consistency constraints. Comparisons with state-of-the-art deep learning models and official forecasts show that this method offers significant advantages in terms of both accuracy and cost-effectiveness, with notable performance improvements in short-term predictions and complex scenarios involving anomalously deflected TC tracks.

\section*{Acknowledgments}
All data used in the analysis are available in public repositories.

1) ERA5 data can be download from \url{https://cds.climate.copernicus.eu/cdsapp#!/dataset/reanalysis-era5-single-levels}

2) CMA-BST data can be download from \url{https://tcdata.typhoon.org.cn/zjljsjj.html}

3) GEBCO data can be download from \url{https://www.gebco.net/data-products/gridded-bathymetry-data}

4) CODCv1 data can be download from \url{https://www.casodc.com/data/}



 


\bibliography{main} 

@inproceedings{sadeghian2019sophie,
  title={{SoPhie}: An attentive {GAN} for predicting paths compliant to social and physical constraints},
  author={Sadeghian, Amir and Kosaraju, Vineet and Sadeghian, Ali and Hirose, Noriaki and Rezatofighi, Hamid and Savarese, Silvio},
  booktitle={Proceedings of the IEEE/CVF Conference on Computer Vision and Pattern Recognition},
  pages={1349--1358},
  year={2019},
  organization={IEEE}
}

@inproceedings{moradi2016sparse,
  title={A sparse recurrent neural network for trajectory prediction of Atlantic hurricanes},
  author={Moradi Kordmahalleh, Majid and Gorji Sefidmazgi, Maryam and Homaifar, Abdollah},
  booktitle={Proceedings of the Genetic and Evolutionary Computation Conference 2016},
  pages={957--964},
  year={2016},
  doi={10.1145/2908812.2908834},
  organization={ACM}
}

@article{ruttgers2019prediction,
  title={Prediction of a typhoon track using a generative adversarial network and satellite images},
  author={R{\"u}ttgers, Marco and Lee, Seungjun and Jeon, Seokhyun and You, Donghyun},
  journal={Scientific Reports},
  volume={9},
  number={1},
  pages={1--15},
  year={2019},
  publisher={Nature Publishing Group}
}

@article{huang2022mmstn,
  title={{MMSTN}: A multi-modal spatial-temporal network for tropical cyclone short-term prediction},
  author={Huang, Cheng and Bai, Cong and Chan, Sixian and Zhang, Jinglin},
  journal={Geophysical Research Letters},
  volume={49},
  number={4},
  pages={e2021GL096898},
  year={2022},
  doi={10.1029/2021GL096898},
  publisher={Wiley}
}

@article{chenFuXiCascadeMachine2023,
  title = {{{FuXi}}: A Cascade Machine Learning Forecasting System for 15-Day Global Weather Forecast},
  shorttitle = {{{FuXi}}},
  author = {Chen, Lei and Zhong, Xiaohui and Zhang, Feng and Cheng, Yuan and Xu, Yinghui and Qi, Yuan and Li, Hao},
  year = {2023},
  month = nov,
  journal = {npj Clim Atmos Sci},
  volume = {6},
  number = {1},
  pages = {190},
  issn = {2397-3722},
  doi = {10.1038/s41612-023-00512-1},
  urldate = {2025-04-22},
  langid = {english}
}

@misc{dendorferMGGANMultiGeneratorModel2021,
  title = {{{MG-GAN}}: {{A Multi-Generator Model Preventing Out-of-Distribution Samples}} in {{Pedestrian Trajectory Prediction}}},
  shorttitle = {{{MG-GAN}}},
  author = {Dendorfer, Patrick and Elflein, Sven and {Leal-Taix{\'e}}, Laura},
  year = {2021},
  month = aug,
  number = {arXiv:2108.09274},
  eprint = {2108.09274},
  primaryclass = {cs},
  publisher = {arXiv},
  doi = {10.48550/arXiv.2108.09274},
  urldate = {2025-04-22},
  archiveprefix = {arxiv},
  langid = {english}
}

@article{fuLongtermTrendTropical2025,
  title = {Long-Term Trend of the Tropical Cyclone Translation Speed over the Western {{North Pacific}} with Track Change},
  author = {Fu, Kai-Jie and Guo, Yi-Peng and Tan, Zhe-Min},
  year = {2025},
  month = jan,
  journal = {Clim Dyn},
  volume = {63},
  number = {1},
  pages = {45},
  issn = {0930-7575, 1432-0894},
  doi = {10.1007/s00382-024-07510-w},
  urldate = {2025-04-22},
  langid = {english}
}

@article{hoangMGANTRAININGGENERATIVE2018,
  title = {{{MGAN}}: {{Training generative adversarial nets with multiple generators}}},
  author = {Hoang, Quan and Nguyen, Tu Dinh and Le, Trung and Phung, Dinh},
  year = {2018},
  langid = {english}
}

@article{huangMGTCFMultiGeneratorTropical2023,
  title = {{{MGTCF}}: {{Multi-Generator Tropical Cyclone Forecasting}} with {{Heterogeneous Meteorological Data}}},
  shorttitle = {{{MGTCF}}},
  author = {Huang, Cheng and Bai, Cong and Chan, Sixian and Zhang, Jinglin and Wu, YuQuan},
  year = {2023},
  month = jun,
  journal = {AAAI},
  volume = {37},
  number = {4},
  pages = {5096--5104},
  issn = {2374-3468, 2159-5399},
  doi = {10.1609/aaai.v37i4.25638},
  urldate = {2025-04-22},
  langid = {english}
}

@article{kangNorthEquatorialCurrent2024,
  title = {The {{North Equatorial Current}} and Rapid Intensification of Super Typhoons},
  author = {Kang, Sok Kuh and Kim, Sung-Hun and Lin, I.-I. and Park, Young-Hyang and Choi, Yumi and Ginis, Isaac and Cione, Joseph and Shin, Ji Yun and Kim, Eun Jin and Kim, Kyeong Ok and Kang, Hyoun Woo and Park, Jae-Hyoung and Bidlot, Jean-Raymond and Ward, Brian},
  year = {2024},
  month = mar,
  journal = {Nat Commun},
  volume = {15},
  number = {1},
  pages = {1742},
  issn = {2041-1723},
  doi = {10.1038/s41467-024-45685-2},
  urldate = {2025-04-22},
  langid = {english}
}

@article{ripple2024StateClimate2024,
  title = {The 2024 State of the Climate Report: {{Perilous}} Times on Planet {{Earth}}},
  shorttitle = {The 2024 State of the Climate Report},
  author = {Ripple, William J and Wolf, Christopher and Gregg, Jillian W and Rockstr{\"o}m, Johan and Mann, Michael E and Oreskes, Naomi and Lenton, Timothy M and Rahmstorf, Stefan and Newsome, Thomas M and Xu, Chi and Svenning, Jens-Christian and Pereira, C{\'a}ssio Cardoso and Law, Beverly E and Crowther, Thomas W},
  year = {2024},
  month = dec,
  journal = {BioScience},
  volume = {74},
  number = {12},
  pages = {812--824},
  issn = {0006-3568, 1525-3244},
  doi = {10.1093/biosci/biae087},
  urldate = {2025-04-22},
  copyright = {https://academic.oup.com/journals/pages/open\_access/funder\_policies/chorus/standard\_publication\_model},
  langid = {english}
}

@article{wangAdvancingForecastingCapabilities2025,
  title = {Advancing Forecasting Capabilities: {{A}} Contrastive Learning Model for Forecasting Tropical Cyclone Rapid Intensification},
  shorttitle = {Advancing Forecasting Capabilities},
  author = {Wang, Chong and Yang, Nan and Li, Xiaofeng},
  year = {2025},
  month = jan,
  journal = {Proc. Natl. Acad. Sci. U.S.A.},
  volume = {122},
  number = {4},
  pages = {e2415501122},
  issn = {0027-8424, 1091-6490},
  doi = {10.1073/pnas.2415501122},
  urldate = {2025-04-22},
  langid = {english}
}

@article{terhaar2025record,
  title = {Record sea surface temperature jump in 2023--2024 unlikely but not unexpected},
  author = {Terhaar, Jens and Burger, Friedrich A. and Vogt, Linus and Fr{\"o}licher, Thomas L. and Stocker, Thomas F.},
  journal = {Nature},
  year = {2025},
  doi = {10.1038/s41586-025-08674-z},
  note = {DOI placeholder: replace with actual DOI when available}
}

@article{globalmarine2025,
  title = {Global marine heatwave of 2023--24 was viewed as unlikely but not impossible given current warming},
  journal = {Nature},
  author={Terhaar, Jens and Burger, Friedrich A},
  year = {2025},
  url = {https://doi.org/10.1038/d41586-025-00888-5}
}

@article{lam2023learning,
  title={Learning skillful medium-range global weather forecasting},
  author={Lam, Remi and Sanchez-Gonzalez, Alvaro and Willson, Matthew and Wirnsberger, Peter and Fortunato, Meire and Alet, Ferran and Ravuri, Suman and Ewalds, Timo and Eaton-Rosen, Zach and Hu, Weihua and others},
  journal={Science},
  volume={382},
  number={6677},
  pages={1416--1421},
  year={2023},
  publisher={American Association for the Advancement of Science}
}

@article{marcos2025global,
  title={Global warming drives a threefold increase in persistence and 1° C rise in intensity of marine heatwaves},
  author={Marcos, Marta and Amores, Angel and Agulles, Miguel and Robson, Jon and Feng, Xiangbo},
  journal={Proceedings of the National Academy of Sciences},
  volume={122},
  number={16},
  pages={e2413505122},
  year={2025},
  publisher={National Academy of Sciences}
}

@article{gao2018nmpt,
  title={A nowcasting model for the prediction of typhoon tracks based on a long short term memory neural network},
  author={Gao, Song and Zhao, Peng and Pan, Bin and Li, Yaru and Zhou, Min and Xu, Jiangling and Zhong, Shan and Shi, Zhenwei},
  journal={Acta Oceanologica Sinica},
  volume={37},
  pages={8--12},
  year={2018},
  publisher={Springer}
}

@article{pan2019DLM,
  title={Tropical cyclone intensity prediction based on recurrent neural networks},
  author={Pan, Bin and Xu, Xia and Shi, Zhenwei},
  journal={Electronics Letters},
  volume={55},
  number={7},
  pages={413--415},
  year={2019},
  publisher={Wiley Online Library}
}

@inproceedings{gupta2018SGAN,
  title={Social gan: Socially acceptable trajectories with generative adversarial networks},
  author={Gupta, Agrim and Johnson, Justin and Fei-Fei, Li and Savarese, Silvio and Alahi, Alexandre},
  booktitle={Proceedings of the IEEE conference on computer vision and pattern recognition},
  pages={2255--2264},
  year={2018}
}

@inproceedings{alemany2019GBRNN,
  title={Predicting hurricane trajectories using a recurrent neural network},
  author={Alemany, Sheila and Beltran, Jonathan and Perez, Adrian and Ganzfried, Sam},
  booktitle={Proceedings of the AAAI conference on artificial intelligence},
  volume={33},
  number={01},
  pages={468--475},
  year={2019}
}

@article{giffard2020FFN,
  title={Tropical cyclone track forecasting using fused deep learning from aligned reanalysis data},
  author={Giffard-Roisin, Sophie and Yang, Mo and Charpiat, Guillaume and Kumler Bonfanti, Christina and K{\'e}gl, Bal{\'a}zs and Monteleoni, Claire},
  journal={Frontiers in big Data},
  volume={3},
  pages={1},
  year={2020},
  publisher={Frontiers Media SA}
}

@article{CMO, 
  title={Typhoon network of Central Meteorological Observatory.},
  author={C.M.O},
  year={2019},
  journal={http://typhoon.nmc.cn/web.html.}
}

@article{jin2024towards,
  title={Towards Robust Tropical Cyclone Wind Radii Estimation with Multi-Modality Fusion and Missing-Modality Distillation},
  author={Jin, Yongjun and Liu, Jia and Ren, Kaijun and Wang, Xiang and Deng, Kefeng and Fan, Zhiqiang and Deng, Chongjiu and Yue, Yinlei},
  journal={IEEE Transactions on Geoscience and Remote Sensing},
  year={2024},
  publisher={IEEE}
}

@article{hu2020comparing,
  title={Comparing the thermal structures of tropical cyclones derived from suomi NPP ATMS and FY-3D microwave sounders},
  author={Hu, Hao and Han, Yang},
  journal={IEEE Transactions on Geoscience and Remote Sensing},
  volume={59},
  number={10},
  pages={8073--8083},
  year={2020},
  publisher={IEEE}
}

@article{chen2020novel,
  title={A novel tensor network for tropical cyclone intensity estimation},
  author={Chen, Zhao and Yu, Xingxing},
  journal={IEEE Transactions on Geoscience and Remote Sensing},
  volume={59},
  number={4},
  pages={3226--3243},
  year={2020},
  publisher={IEEE}
}

@article{zhang2019tropical,
  title={Tropical cyclone intensity estimation using two-branch convolutional neural network from infrared and water vapor images},
  author={Zhang, Rui and Liu, Qingshan and Hang, Renlong},
  journal={IEEE Transactions on Geoscience and Remote Sensing},
  volume={58},
  number={1},
  pages={586--597},
  year={2019},
  publisher={IEEE}
}

@article{ma2024multiscale,
  title={A multiscale and multilayer feature extraction network with dual attention for tropical cyclone intensity estimation},
  author={Ma, Zhaoyang and Yan, Yunfeng and Lin, Jianmin and Ma, Dongfang},
  journal={IEEE Transactions on Geoscience and Remote Sensing},
  volume={62},
  pages={1--15},
  year={2024},
  publisher={IEEE}
}

@article{wang2021tropical,
  title={Tropical cyclone intensity estimation from geostationary satellite imagery using deep convolutional neural networks},
  author={Wang, Chong and Zheng, Gang and Li, Xiaofeng and Xu, Qing and Liu, Bin and Zhang, Jun},
  journal={IEEE Transactions on Geoscience and Remote Sensing},
  volume={60},
  pages={1--16},
  year={2021},
  publisher={IEEE}
}

@article{tian2021estimation,
  title={Estimation of tropical cyclone intensity using synthetic satellite microwave temperature anomaly structure and a multifeature distribution learning network},
  author={Tian, Miao and Zhang, Taidong and Liu, Guanghui and Lin, Lin},
  journal={IEEE Transactions on Geoscience and Remote Sensing},
  volume={60},
  pages={1--12},
  year={2021},
  publisher={IEEE}
}

@article{huang2025benchmark,
  title={Benchmark dataset and deep learning method for global tropical cyclone forecasting},
  author={Huang, Cheng and Mu, Pan and Zhang, Jinglin and Chan, Sixian and Zhang, Shiqi and Yan, Hanting and Chen, Shengyong and Bai, Cong},
  journal={Nature Communications},
  volume={16},
  number={1},
  pages={5923},
  year={2025},
  publisher={Nature Publishing Group UK London}
}

@article{ma2025philippine,
  title={Philippine archipelago and South China Sea monsoon plus ocean cooling buffer Northwestern Pacific super typhoons},
  author={Ma, Tian and Yu, Wei-Dong and Speich, Sabrina and Zhao, Hai-Kun and Xin, Rui and Luo, Hao and Wu, Li-Guang},
  journal={Nature Communications},
  volume={16},
  number={1},
  pages={7395},
  year={2025},
  publisher={Nature Publishing Group UK London}
}

@inproceedings{chowdhury2023patch,
  title={Patch-level routing in mixture-of-experts is provably sample-efficient for convolutional neural networks},
  author={Chowdhury, Mohammed Nowaz Rabbani and Zhang, Shuai and Wang, Meng and Liu, Sijia and Chen, Pin-Yu},
  booktitle={International Conference on Machine Learning},
  pages={6074--6114},
  year={2023},
  organization={PMLR}
}

@article{bi2023accurate,
  title={Accurate medium-range global weather forecasting with 3D neural networks},
  author={Bi, Kaifeng and Xie, Lingxi and Zhang, Hengheng and Chen, Xin and Gu, Xiaotao and Tian, Qi},
  journal={Nature},
  volume={619},
  number={7970},
  pages={533--538},
  year={2023},
  publisher={Nature Publishing Group UK London}
}

@article{subich2025fixing,
  title={Fixing the double penalty in data-driven weather forecasting through a modified spherical harmonic loss function},
  author={Subich, Christopher and Husain, Syed Zahid and Separovic, Leo and Yang, Jing},
  journal={arXiv preprint arXiv:2501.19374},
  year={2025}
}

@inproceedings{alahi2016social,
  title={Social lstm: Human trajectory prediction in crowded spaces},
  author={Alahi, Alexandre and Goel, Kratarth and Ramanathan, Vignesh and Robicquet, Alexandre and Fei-Fei, Li and Savarese, Silvio},
  booktitle={Proceedings of the IEEE conference on computer vision and pattern recognition},
  pages={961--971},
  year={2016}
}

@article{kosaraju2019social,
  title={Social-bigat: Multimodal trajectory forecasting using bicycle-gan and graph attention networks},
  author={Kosaraju, Vineet and Sadeghian, Amir and Mart{\'\i}n-Mart{\'\i}n, Roberto and Reid, Ian and Rezatofighi, Hamid and Savarese, Silvio},
  journal={Advances in neural information processing systems},
  volume={32},
  year={2019}
}

@inproceedings{amirian2019social,
  title={Social ways: Learning multi-modal distributions of pedestrian trajectories with gans},
  author={Amirian, Javad and Hayet, Jean-Bernard and Pettr{\'e}, Julien},
  booktitle={Proceedings of the IEEE/CVF Conference on Computer Vision and Pattern Recognition Workshops},
  pages={0--0},
  year={2019}
}

@inproceedings{dendorfer2020goal,
  title={Goal-gan: Multimodal trajectory prediction based on goal position estimation},
  author={Dendorfer, Patrick and Osep, Aljosa and Leal-Taix{\'e}, Laura},
  booktitle={Proceedings of the Asian Conference on Computer Vision},
  year={2020}
}

@inproceedings{fernando2018gd,
  title={Gd-gan: Generative adversarial networks for trajectory prediction and group detection in crowds},
  author={Fernando, Tharindu and Denman, Simon and Sridharan, Sridha and Fookes, Clinton},
  booktitle={Asian conference on computer vision},
  pages={314--330},
  year={2018},
  organization={Springer}
}

@article{goodfellow2014generative,
  title={Generative adversarial nets},
  author={Goodfellow, Ian J and Pouget-Abadie, Jean and Mirza, Mehdi and Xu, Bing and Warde-Farley, David and Ozair, Sherjil and Courville, Aaron and Bengio, Yoshua},
  journal={Advances in neural information processing systems},
  volume={27},
  year={2014}
}

@article{chen2022towards,
  title={Towards understanding mixture of experts in deep learning},
  author={Chen, Zixiang and Deng, Yihe and Wu, Yue and Gu, Quanquan and Li, Yuanzhi},
  journal={arXiv preprint arXiv:2208.02813},
  year={2022}
}

@article{zhong2022meta,
  title={Meta-dmoe: Adapting to domain shift by meta-distillation from mixture-of-experts},
  author={Zhong, Tao and Chi, Zhixiang and Gu, Li and Wang, Yang and Yu, Yuanhao and Tang, Jin},
  journal={Advances in Neural Information Processing Systems},
  volume={35},
  pages={22243--22257},
  year={2022}
}

@article{jacobs1991textordfeminineadaptive,
  title={Adaptive Mixtures of Local Experts},
  author={Jacobs, RA and Jordan, MI and Nowlan, SJ and Hinton, GE},
  journal={Neural Computation},
  volume={3},
  year={1991}
}

@article{shazeer2017outrageously,
  title={Outrageously large neural networks: The sparsely-gated mixture-of-experts layer},
  author={Shazeer, Noam and Mirhoseini, Azalia and Maziarz, Krzysztof and Davis, Andy and Le, Quoc and Hinton, Geoffrey and Dean, Jeff},
  journal={arXiv preprint arXiv:1701.06538},
  year={2017}
}

@ARTICLE{ZhuAttention2025,
  author={Zhu, Peiyuan and Zhao, Shengjie and Deng, Hao and Han, Fengxia},
  journal={IEEE Transactions on Intelligent Transportation Systems}, 
  title={Attentive Radiate Graph for Pedestrian Trajectory Prediction in Disconnected Manifolds}, 
  year={2025},
  volume={26},
  number={6},
  pages={7755-7769},
  doi={10.1109/TITS.2025.3555390}}

@article{knapp2017hursat,
  title={The international best track archive for climate stewardship (IBTrACS) unifying tropical cyclone data},
  author={Knapp, Kenneth R and Kruk, Michael C and Levinson, David H and Diamond, Howard J and Neumann, Charles J},
  journal={Bulletin of the American Meteorological Society},
  volume={91},
  number={3},
  pages={363--376},
  year={2010},
  publisher={American Meteorological Society}
}

@inproceedings{chen2018tcir,
  title={Rotation-blended CNNs on a new open dataset for tropical cyclone image-to-intensity regression},
  author={Chen, Boyo and Chen, Buo-Fu and Lin, Hsuan-Tien},
  booktitle={Proceedings of the 24th ACM SIGKDD international conference on knowledge discovery \& data mining},
  pages={90--99},
  year={2018}
}

@article{kitamoto2023digital,
  title={Digital typhoon: Long-term satellite image dataset for the spatio-temporal modeling of tropical cyclones},
  author={Kitamoto, Asanobu and Hwang, Jared and Vuillod, Bastien and Gautier, Lucas and Tian, Yingtao and Clanuwat, Tarin},
  journal={Advances in Neural Information Processing Systems},
  volume={36},
  pages={40623--40636},
  year={2023}
}

@inproceedings{wang2025global,
  title={Global tropical cyclone intensity forecasting with multi-modal multi-scale causal autoregressive model},
  author={Wang, Xinyu and Chen, Kang and Liu, Lei and Han, Tao and Li, Bin and Bai, Lei},
  booktitle={ICASSP 2025-2025 IEEE International Conference on Acoustics, Speech and Signal Processing (ICASSP)},
  pages={1--5},
  year={2025},
  organization={IEEE}
}

@article{baiman2026watch,
  title={Watch an AI Weather Model Learn (and Unlearn) Tropical Cyclones},
  author={Baiman, Rebecca and Mahesh, Ankur and Barnes, Elizabeth A},
  journal={arXiv preprint arXiv:2603.20541},
  year={2026}
}

@article{zhu2024moe,
  title={Moe jetpack: From dense checkpoints to adaptive mixture of experts for vision tasks},
  author={Zhu, Xingkui and Guan, Yiran and Liang, Dingkang and Chen, Yuchao and Liu, Yuliang and Bai, Xiang},
  journal={Advances in Neural Information Processing Systems},
  volume={37},
  pages={12094--12118},
  year={2024}
}

@article{zhu2023sira,
  title={Sira: Sparse mixture of low rank adaptation},
  author={Zhu, Yun and Wichers, Nevan and Lin, Chu-Cheng and Wang, Xinyi and Chen, Tianlong and Shu, Lei and Lu, Han and Liu, Canoee and Luo, Liangchen and Chen, Jindong and others},
  journal={arXiv preprint arXiv:2311.09179},
  year={2023}
}

@article{li2023merge,
  title={Merge, then compress: Demystify efficient smoe with hints from its routing policy},
  author={Li, Pingzhi and Zhang, Zhenyu and Yadav, Prateek and Sung, Yi-Lin and Cheng, Yu and Bansal, Mohit and Chen, Tianlong},
  journal={arXiv preprint arXiv:2310.01334},
  year={2023}
}

@inproceedings{chen2023adamv,
  title={Adamv-moe: Adaptive multi-task vision mixture-of-experts},
  author={Chen, Tianlong and Chen, Xuxi and Du, Xianzhi and Rashwan, Abdullah and Yang, Fan and Chen, Huizhong and Wang, Zhangyang and Li, Yeqing},
  booktitle={proceedings of the IEEE/CVF international conference on computer vision},
  pages={17346--17357},
  year={2023}
}

@article{ma2026interactions,
  title={Interactions of tropical cyclones with global energy and water cycles},
  author={Ma, Zhanhong and Cheng, Lijing and Camargo, Suzana J and Trenberth, Kevin E and Lin, II and Foltz, Gregory R and Chavas, Daniel R and Zhang, Deyuan and Ritchie, Elizabeth A and Fei, Jianfang and others},
  journal={Nature Reviews Earth \& Environment},
  pages={1--19},
  year={2026},
  publisher={Nature Publishing Group UK London}
}











\newpage

\vspace{11pt}


\vspace{11pt}

\vfill

\end{document}